\documentclass[journal,twoside,web]{IEEETran}
\usepackage{orcidlink}
\hypersetup{
    colorlinks,
    linkcolor={black!50!black},
    citecolor={blue!50!black},
    urlcolor={black!80!black}
}
\usepackage{tmi}
\usepackage{cite}
\usepackage{amsmath,amssymb,amsfonts}
\usepackage{algorithmic}
\usepackage{graphicx}
\usepackage{textcomp}
\usepackage{multirow}
\usepackage{algorithm}
\usepackage{algorithmic}
\usepackage{flushend}
\usepackage{lettrine}
\usepackage{arydshln}

\makeatletter
\newcommand*\titleheader[1]{\gdef\@titleheader{#1}}
\AtBeginDocument{%
  \let\st@red@title\@title
  \def\@title{%
    \bgroup\normalfont\large\centering\@titleheader\par\egroup
    \vskip1.5em\st@red@title}
}
\makeatother

\def\BibTeX{{\rm B\kern-.05em{\sc i\kern-.025em b}\kern-.08em
    T\kern-.1667em\lower.7ex\hbox{E}\kern-.125emX}}
\title{QTSeg: A Query Token-Based Dual-Mix Attention Framework with Multi-Level Feature Distribution for Medical Image Segmentation}
\titleheader{\small{This work has been submitted to the IEEE for possible publication. Copyright may be transferred without notice, after which this version may no longer be accessible.}}
\author{Phuong-Nam Tran$^1$, Nhat Truong Pham$^2$, Duc Ngoc Minh Dang$^3$, Eui-Nam Huh$^4$ and Choong Seon Hong$^{4,*}$ \\
\small{
$^1$ Department of Artificial Intelligence Kyung Hee University, Yongin, Republic of Korea\\
$^2$ Department of Integrative Biotechnology Sungkyunkwan University, Suwon, Gyeonggi-do, Republic of Korea\\
$^3$ Department of Computing Fundamental, FPT University, Ho Chi Minh Campus, Vietnam\\
$^4$ Department of Computer Science and Engineering Kyung Hee University, Yongin, Republic of Korea\\
tpnam0901@khu.ac.kr, truongpham96@skku.edu, ducdnm2@fe.edu.vn, \{johnhuh,cshong\}@khu.ac.kr
}
\thanks{* Corresponding author.}
}

\begin{document}

\maketitle

\begin{abstract}
Medical image segmentation plays a crucial role in assisting healthcare professionals with accurate diagnoses and enabling automated diagnostic processes. Traditional convolutional neural networks (CNNs) often struggle with capturing long-range dependencies, while transformer-based architectures, despite their effectiveness, come with increased computational complexity. Recent efforts have focused on combining CNNs and transformers to balance performance and efficiency, but existing approaches still face challenges in achieving high segmentation accuracy while maintaining low computational costs. Furthermore, many methods underutilize the CNN encoder's capability to capture local spatial information, concentrating primarily on mitigating long-range dependency issues. To address these limitations, we propose QTSeg, a novel architecture for medical image segmentation that effectively integrates local and global information. QTSeg features a dual-mix attention decoder designed to enhance segmentation performance through: (1) a cross-attention mechanism for improved feature alignment, (2) a spatial attention module to capture long-range dependencies, and (3) a channel attention block to learn inter-channel relationships. Additionally, we introduce a multi-level feature distribution module, which adaptively balances feature propagation between the encoder and decoder, further boosting performance. Extensive experiments on five publicly available datasets covering diverse segmentation tasks, including lesion, polyp, breast cancer, cell, and retinal vessel segmentation, demonstrate that QTSeg outperforms state-of-the-art methods across multiple evaluation metrics while maintaining lower computational costs.

% These instructions provide guidelines for preparing papers for IEEE Transactions,
% but this version is specifically written to describe submission to IEEE TMI.
% Use this document as a template if you are using \LaTeX.
% Otherwise, use this document as an instruction set.
% The electronic file of your paper will be formatted further at IEEE.
% Paper titles should be written in uppercase and lowercase letters, not all uppercase.
% Avoid writing long formulas with subscripts in the title;
% short formulas that identify the elements are fine (e.g., "Nd--Fe--B").
% Keep the title short and do not write ``(Invited)'' in the title.
% Full names of authors are preferred in the author field, but are not required.
% Put a space between authors' initials. Only authors may appear in the author line
% of a manuscript. Authors are defined as individuals who have made an identifiable
% intellectual contribution to a manuscript to the extent that the individual can defend its contents.
% Define all symbols used in the abstract. Do not cite references in the abstract.
% Keep the abstract to 250 words or less.
\end{abstract}

\begin{IEEEkeywords}
% Enter about five key words or phrases in alphabetical order, separated by commas.
Convolutional neural networks, medical image segmentation, self-attention mechanism, transformer, dual-mix attention.
\end{IEEEkeywords}

\section{Introduction}

\begin{figure}[!ht]
    \centering
    \includegraphics[width=\linewidth]{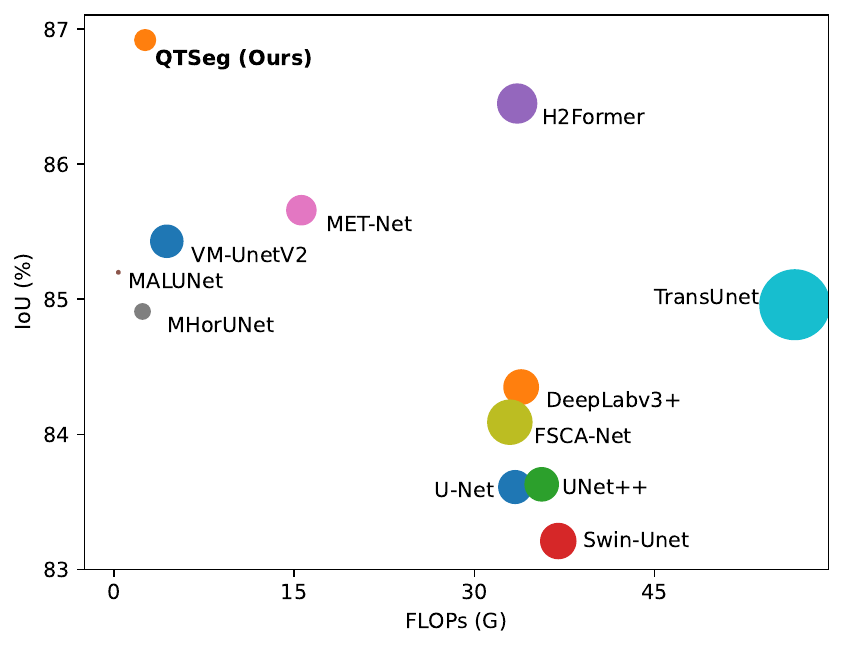}
    \caption{The comparison of Dice score and FLOPs on the ISIC2016 dataset between QTSeg and other methods. It shows that our proposed QTSeg outperforms all the other methods in IoU score with small FLOPs. Larger circles indicate higher parameter sizes.}
    \label{fig:acc_flops}
\end{figure}

% Medical application
\IEEEPARstart{M}{edical} image segmentation is a fundamental task in medical imaging, serving as a cornerstone for precise diagnosis, treatment planning, and quantitative analysis across various clinical applications~\cite{xuyan2024advancesMIS}.
By partitioning an image into distinct regions, segmentation enables healthcare professionals to analyze anatomical structures and pathological conditions in detail.
This technique is crucial for tasks such as tumor identification, organ volume measurement, and surgical planning. 
Additionally, it plays a pivotal role in training AI models for automated diagnosis~\cite{rahman2023MISviaCAD}, reducing the workload of medical professionals while enhancing diagnostic accuracy.
Given its significance, the development of automated and highly accurate medical image segmentation methods remains an essential and pressing need.

With rapid advancements in artificial intelligence, computer vision has made significant strides in image segmentation.
However, a substantial gap still exists between medical and natural images due to differences in structure, texture, and complexity~\cite{morra2021icpr}.
Additionally, the interpretation of medical images requires specialized expertise, resulting in a scarcity of labeled training data.
To address these challenges, researchers have explored various strategies, including transfer learning~\cite{kimhee2022transfer}, domain adaptation~\cite{guan2022ieeetbe}, and hybrid models~\cite{h2former, missformer}. 
Transfer learning fine-tune models pretrained on natural images for medical image segmentation, while domain adaptation adjusts models to better handle the unique characteristics of medical images.
Hybrid models combine traditional segmentation techniques with deep learning methods and novel architectures to improve performance with limited datasets. 

Building on these advancements, convolutional neural networks (CNNs) have been widely adopted for medical image segmentation, with models such as U-Net~\cite{unet}, UNet++~\cite{unet++}, and nnUNet~\cite{isensee2021nnunet} demonstrating significant success. 
These models automatically learn and extract features from medical images, aiding in the diagnosis and treatment planning of conditions such as brain tumors, lung diseases, and other anomalies.
Despite their effectiveness, CNNs struggle to capture long-range dependencies in medical images~\cite{rahman2024mist}. 
This limitation arises because CNNs primarily focus on local features, making it difficult to incorporate contextual information across larger spatial areas.
To address this challenge, researchers have explored transformer architectures\cite{vaswani2017attention}, originally developed for natural language processing, due to their strong capability in handling long-range dependencies\cite{kelei2023transformerMIA}. 
Transformer-based models such as TransUNet~\cite{transunet} and SwinUnet~\cite{swinunet} have been adapted for medical imaging tasks, achieving state-of-the-art results. 
By either combining CNNs with transformers or leveraging pure transformer architectures, these models have significantly improved segmentation performance in medical image analysis.

However, while transformers excel at capturing long-range dependencies, they often struggle with preserving local structural details, which are crucial for accurate segmentation.
This limitation can lead to suboptimal performance, particularly in tasks requiring precise localization~\cite{honglin2024cmlformer}. 
Additionally, transformers are computationally demanding and require large-scale annotated datasets~\cite{xian2023lighterbetter}, which poses challenges in the medical domain where labeled data is often scarce.

% Despite these obstacles, current research endeavours focus on addressing these challenges by integrating transformers with other novel architecture to capitalize on their individual strengths and enhance overall performance in medical image analysis. 
% One promising direction is the development of hybrid transformer architectures, such as MISSFormer~\cite{missformer} and H2Former~\cite{nnformer}, which aim to combine the strengths of the transformer and CNN. 
To address these issues, current research efforts focus on hybrid transformer architectures, such as MISSFormer\cite{missformer}, which integrate CNN into transformers to leverage their respective strengths. 
However, these models often rely primarily on token-wise self-attention, neglecting channel-wise interactions and multi-scale feature learning, which are essential for capturing intricate anatomical structures~\cite{h2former}. 
This limitation can hinder their ability to fully capture the intricate details and structural priors essential for segmentation tasks, leading to suboptimal performance in certain scenarios. 
Moreover, while combining CNNs with transformers enhances performance, it often increases computational complexity, underscoring the need for efficient and lightweight architectures for practical deployment in medical imaging applications~\cite{xian2023lighterbetter}.

Motivated by the limitations of existing approaches, we introduce QTSeg~\footnote{Code is available at https://github.com/tpnam0901/QTSeg (v1.0.0)}, a model designed to effectively handle long-range dependencies at multiple feature levels while balancing accuracy and computational efficiency.
Our approach enhances segmentation performance by integrating a dual-mix attention module (DMA), which captures both local details and long-range dependencies while minimizing computational overhead.
The DMA module consists of four key submodules: channel attention block (CAB), spatial attention module (SAM)~\cite{sanghyun2018cbam}, cross-token feature attention (CTFA), and cross-feature token attention (CFTA). 
While CAB captures inter-channel relationships among features, SAM applies spatial attention to address long-range dependencies inherent in CNN architectures.
Additionally, CTFA and CFTA are utilized to combine information from feature embeddings and query tokens through cross-attention~\cite{chunfu2021cross_attention}.
Drawing inspiration from U-shaped architecture, we have designed a mask decoder with multiple dual-mix attention decoders (DMADs) attached to each feature embedding to learn the relationships between feature embeddings at each stage.
Unlike conventional U-shaped architectures that typically use simple multi-layer perceptrons (MLPs) or a convolution layer for mask extraction, our DMAD module aligns query tokens with feature embeddings across different levels, from low-level to high-level embeddings.
This alignment facilitates a more comprehensive understanding of the relationships between all features in the image.
Experimental results demonstrate the efficiency of our proposed QTSeg method, which achieves a competitive parameter count and low floating-point operations (FLOPs), all while delivering high performance on benchmark datasets, as shown in Fig.~\ref{fig:acc_flops}.

% Main contribution
Our main contribution can be summarized as follows:
\begin{enumerate}
    \item We introduce the QTSeg model, which efficiently addresses long-range dependencies in CNN-based models while maintaining low computational complexity.
    \item We present DMADs, which utilize query tokens to extract target masks from multi-level feature embeddings, enhancing performance through the utilization of cross-attention modules.
    \item We propose the DMA module, designed to learn inter-channel relationships and capture long-range dependencies in features extracted by the encoder.
    \item We conduct extensive experiments demonstrating the effectiveness of our approach, which outperforms existing state-of-the-art models in medical image segmentation.
\end{enumerate}

The structure of the remaining sections in this paper is outlined as follows. Section~\ref{sec:related_work} presents the methodologies employed in previous studies. Section~\ref{sec:method} provides a detailed explanation of our proposed method. Section~\ref{sec:experiments} presents the experimental results obtained from different datasets and highlights the significance of each module in our architecture. Finally, in Section~\ref{sec:conclusion}, we draw conclusions based on the findings presented in this paper.

\section{Related Work}
\label{sec:related_work}
\subsection{Convolutional Neural Networks in Medical Image Segmentation}
In recent years, several CNN architectures have been developed for medical image segmentation, taking advantage of their ability to effectively capture spatial features.
One of the most well-known models is U-Net~\cite{unet}, recognized for its U-shaped architecture, which incorporates skip connections from previous layers to preserve essential information throughout the network.
Building upon the achievements of U-Net, several subsequent models have emerged following a similar architecture, including UNet++~\cite{unet++}, Dense-UNet~\cite{dense_unet}, nnUnet~\cite{isensee2021nnunet}, and Attention Unet~\cite{attention_unet}. 
On the other hand, approaches that use vanilla skip connection blocks, such as DeepMedic~\cite{kamnitsas2017deepmedic} and SegResNet~\cite{myronenko2019segresnet}, were also introduced, delivering comparable performance while maintaining low resource requirements.
These models have significantly advanced both general image segmentation and, more specifically, medical image segmentation, further highlighting the potential of CNN-based approaches in medical computer vision.

\subsection{Transformers in Medical Image Segmentation}
Recently, the transformer architecture has shown great potential in various computer vision tasks~\cite{cv_task1, swintexco, cv_task2, medsam, he2021transreid} by addressing the challenge of long-range dependencies in traditional CNN networks.
A major milestone in this evolution was the emergence of the Vision Transformer (ViT) architecture~\cite{vit}, which integrates attention mechanisms into computer vision. 
By dividing the input image into a sequence of patches and applying attention mechanisms to these features, ViT significantly improves model performance in various computer vision tasks.

Building on this success, the Swin Transformer~\cite{swin_transformer} introduced a window-based approach for implementing self-attention within local windows, reducing computational complexity while enhancing performance. 
With the demonstrated effectiveness of transformers, many studies have combined this architecture with other novel techniques to improve performance in medical image segmentation tasks. 
A common approach involves combining CNN and transformer models to capitalize on their respective strengths. 

One of the early adopters of this hybrid approach is TransUNet~\cite{transunet}, which pioneers the fusion of CNNs and transformers for medical image segmentation.
Rather than replacing the CNN, TransUNet uses the transformer's capabilities to enhance the existing CNN architecture by adding it between the encoder and decoder.
Expanding on this idea, Swin-UNet\cite{swinunet} introduced a pure transformer-based architecture to extract target features in medical image segmentation.
However, it is important to note that due to the quadratic complexity of the transformer, these models still require substantial computational resources to generate outputs.

\subsection{Analysis of Previous Work}
\begin{figure}[ht]
    \centering
    \includegraphics[width=\linewidth, height=3.7cm]{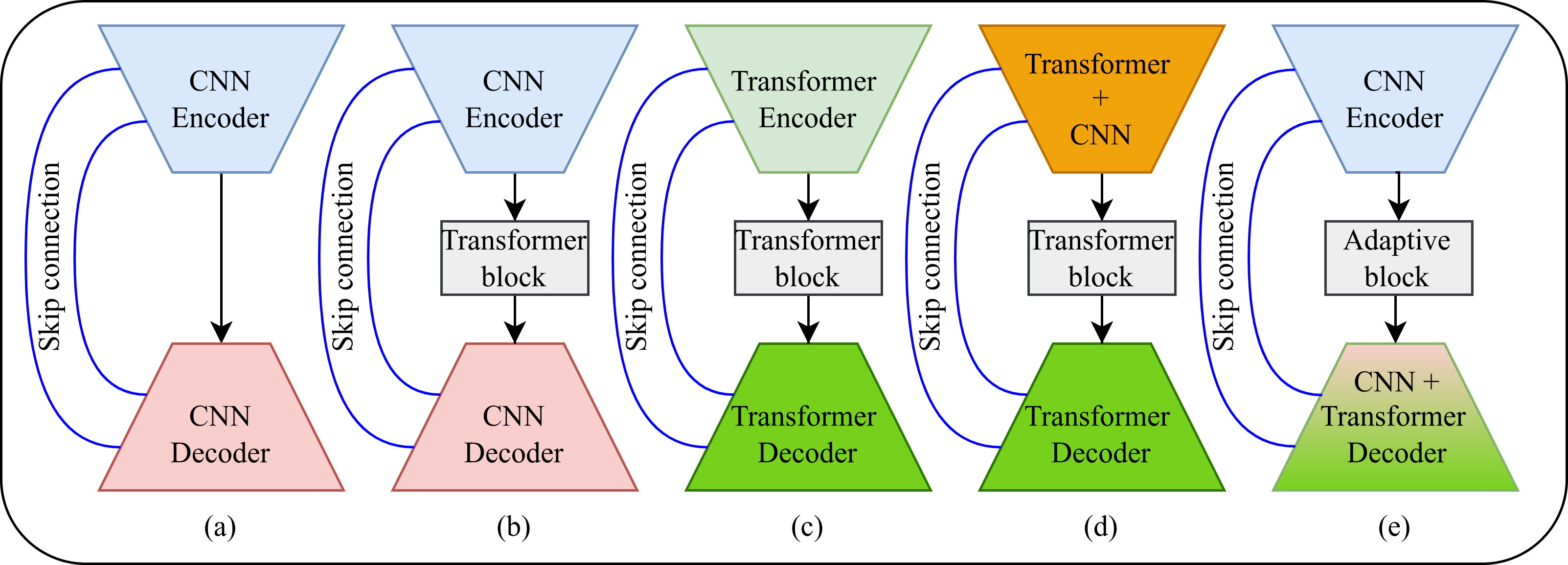}
    \caption{Comparative conceptual of architectures for medical image segmentation. \textcolor{blue}{a)} The vanilla technique using CNN U-Shaped (e.g., UNet~\cite{unet}). \textcolor{blue}{b)} The cascaded architecture of CNN and transformer module utilizing in TransUnet~\cite{transunet}. \textcolor{blue}{c)} The pure transformer architecture for image segmentation in SwinUnet~\cite{swinunet}. \textcolor{blue}{d)} The efficient architecture in encoder image feature using a hybrid transformer in H2Former~\cite{h2former}. \textcolor{blue}{e)} Our proposed QTSeg architecture.}
    \label{fig:conceptual_comparison}
\end{figure}

In recent years, researchers have explored various architectures that combine CNN and transformer components to enhance performance. 
These diverse architectural approaches are summarized in Fig.~\ref{fig:conceptual_comparison}, divided into five approaches. 
For the initial architectures, the CNN U-shaped design is commonly employed due to its simplicity and effectiveness. 
Illustrated in Fig.~\ref{fig:conceptual_comparison}a, this architecture comprises both CNN encoder and decoder with skip connections between them, which is commonly known through U-Net~\cite{unet}.
As illustrated in Fig.~\ref{fig:conceptual_comparison}a, this architecture consists of both CNN encoder and decoder, with skip connections between them, a design widely known through U-Net~\cite{unet}.
To further enhance the performance of this architecture, a transformer block is inserted between the encoder and decoder to learn feature attention at a low level, as shown in Fig.~\ref{fig:conceptual_comparison}b and TransUNet~\cite{transunet}.

An alternative approach involves eliminating the CNN component and utilizing a pure transformer, as shown in Fig.~\ref{fig:conceptual_comparison}c and SwinUNet~\cite{swinunet}. 
However, opting for pure transformers introduces challenges, particularly in terms of computational complexity within computer vision tasks.
To mitigate these issues, researchers have focused on reducing parameters by developing hybrid CNN-transformer encoder-decoder architectures.
As depicted in Fig.~\ref{fig:conceptual_comparison}d, the fusion of CNN and transformer components has shown considerable potential, offering higher accuracy while maintaining competitive parameter requirements, as evidenced in H2Former~\cite{h2former}.
However, these hybrid architectures still face challenges, as they do not fully capitalize on the strengths of both CNNs and transformers while exhibiting high parameter counts and FLOPs.

CNN architectures excel at extracting local spatial information effectively, but they struggle to capture long-range dependencies.
In contrast, transformers are well-suited for capturing long-range dependencies, but their quadratic computational complexity poses a challenge for deploying in low-facility hospitals. 
To leverage the strengths of both CNN and transformer architectures, hybrid models have been developed that combine convolutional layers with attention mechanisms.
However, these hybrid models often come with increased complexity and high FLOPs, making them more resource-intensive than models relying solely on CNNs.
To tackle these challenges, we propose the QTSeg with an adaptive block and DMAD, as shown in Fig.~\ref{fig:conceptual_comparison}e.
This architecture offers the flexibility to interchangeably replace the encoder with either CNN or transformer components, leveraging powerful pre-trained models for scalability and adaptability.
The effectiveness of this innovative approach is demonstrated in Section~\ref{exp:results} and Section~\ref{ablation}.

\begin{figure*}[ht]
    \centering
    \includegraphics[width=\linewidth]{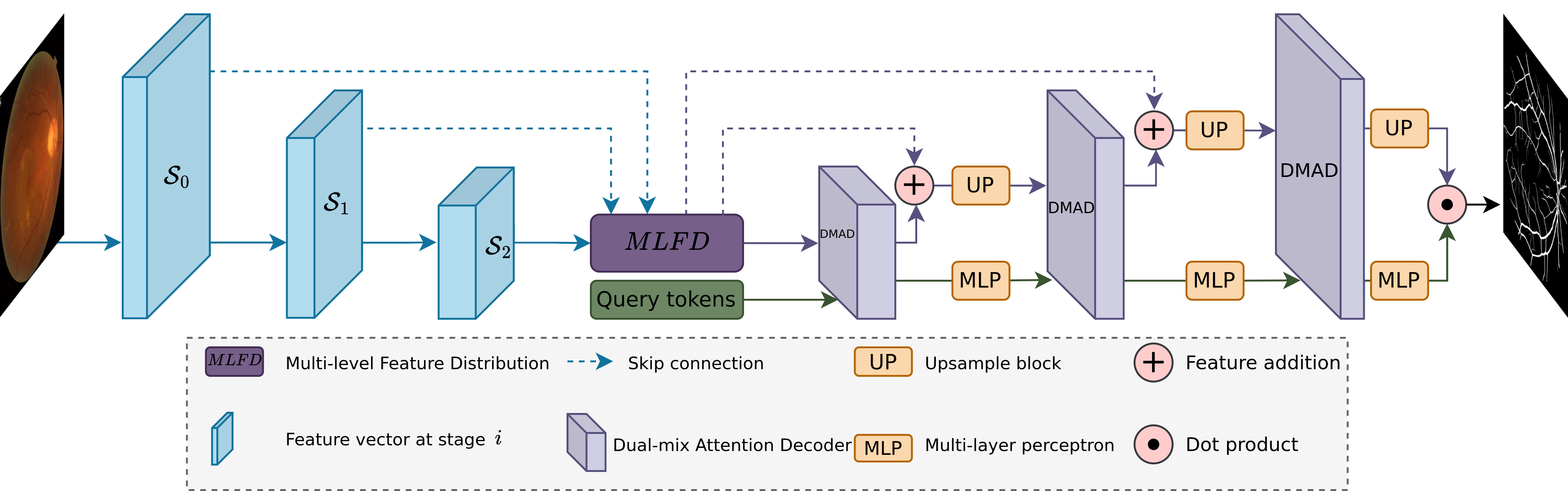}
    \caption{The overall architecture of the proposed method.}
    \label{fig:proposed_method_overview}
\end{figure*}

\section{Methodology}
\label{sec:method}
\subsection{Overall architecture}

The overall pipeline of our proposed method is depicted in Fig.~\ref{fig:proposed_method_overview}. In this framework, the encoder can be any CNN multiple-feature extraction model such as ResNet~\cite{kaiming2016resnet}, YOLOv8~\cite{varghese2024yolov8}. For the decoder, we propose a mask decoder that makes use of multiple DMADs to optimize the utilization of features extracted by the encoder. The decoder follows a U-shaped~\cite{unet} architecture that comprises three main modules: DMA, MLP, and Upsampling blocks. Additionally, a multi-level feature distribution (MLFD) adaption module is incorporated between the encoder and decoder to evenly distribute features across all levels before transmitting them to the decoder. During the decoding phase, query tokens are utilized to align with feature embeddings to enhance mask-predicted performance in the final layer through cross-attention~\cite{chunfu2021cross_attention}. 

\subsection{Mask Decoder}
\begin{figure}[ht]
    \centering
    \includegraphics[width=\linewidth]{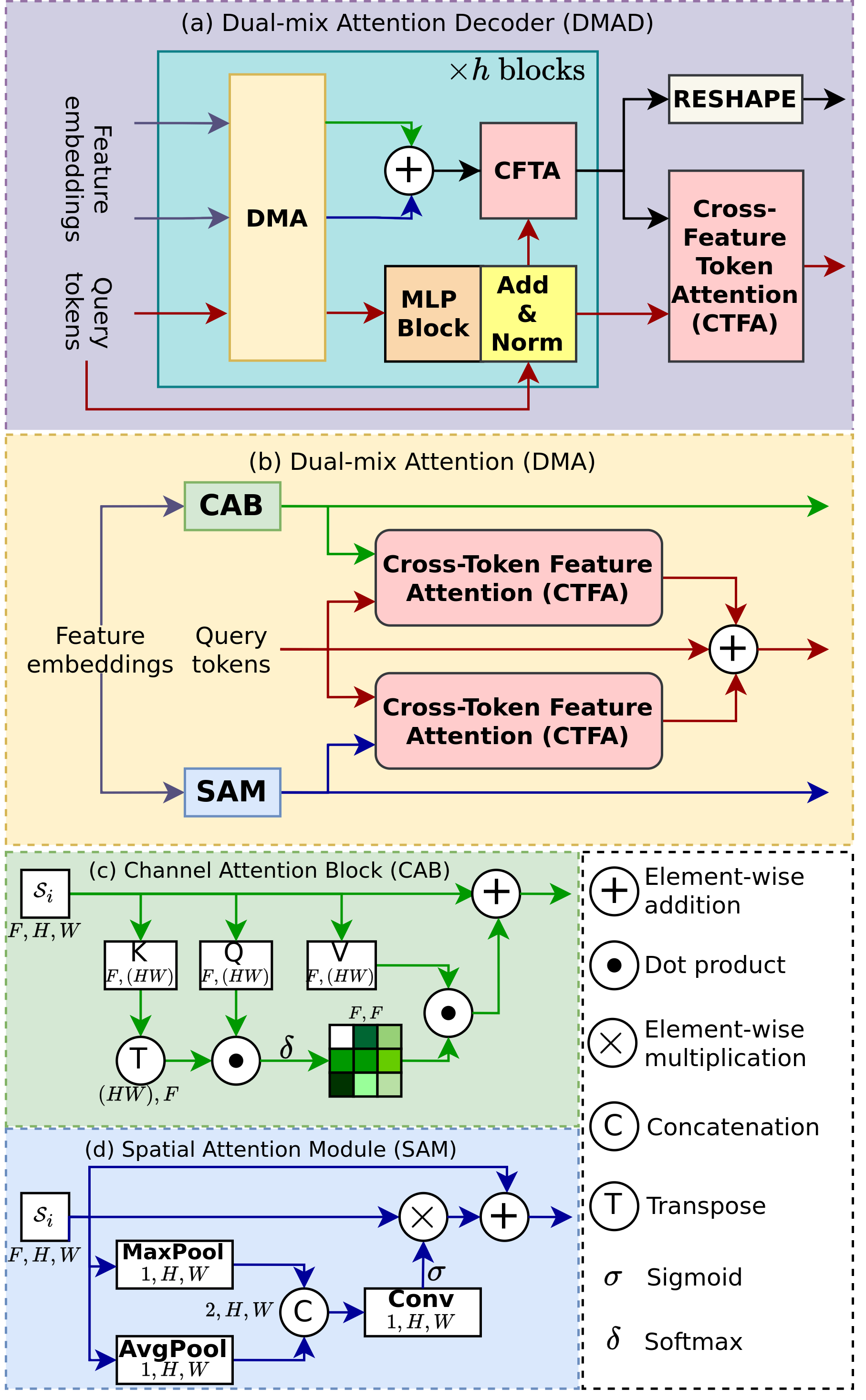}
    \caption{Detailed illustration of the Dual-Mix Attention Decoder.}
    \label{fig:mask_decoder}
\end{figure}

The overall architecture of the decoder is illustrated in Fig.~\ref{fig:proposed_method_overview}, resembling the well-known U-shaped design~\cite{unet}. However, our architecture differs from the conventional approach by incorporating a DMAD and query tokens rather than flipping the previous encoder architecture. Let $QT_i \in \mathbb{R}^{F_i\times N}$ denote the query tokens at the $i^{th}$ stage and  $\mathcal{S}_i \in \mathbb{R}^{F_i\times H_i\times W_i}$ represent the feature extracted by the encoder at stage $i^{th}$, with $F$, $H$, $W$ and $N$ indicating the feature dimension, height, width and the number of class, respectively. Our decoder efficiently extracts features from the encoder through the DMAD presented in Fig.~\ref{fig:mask_decoder}a, leveraging the CAB and SAM~\cite{sanghyun2018cbam}  to capture the inter-channel relationship of features and long-range features. Moreover, the DMAD aligns $\mathcal{S}_i$ with $QT_i$ to enhance mask generation in the final stage through CFTA and CTFA. Following each DMAD, we utilize an Upsampling block to increase the feature embedding size for the next stage prediction. This block comprises a 2D convolutional transpose, layer normalization, and GELU activation. Furthermore, an MLP block is utilized after the DMAD to map $QT_i$ to $QT_{i-1}$, transitioning from the $F_i$ to $F_{i-1}$ dimension. In the final layer, the dot product between the feature embeddings and query tokens is utilized to generate target masks. 

\subsubsection{Dual-mix attention}
\label{subsec:DMADlock}
To effectively capture the inter-channel relationships within the features, we introduce the CAB module, which performs self-attention on the feature channels, as depicted in Fig.~\ref{fig:mask_decoder}c. The CAB module is built upon the attention mechanism, where $F$ is considered the number of tokens and $H\times W$ represents the feature dimension of the tokens. The formulation of the query $Q\in \mathbb{R}^{F\times HW}$, key $K\in \mathbb{R}^{F\times HW}$, and value $V\in \mathbb{R}^{F\times HW}$ of the CAB block can be expressed as follows:

\begin{equation}
    \label{equa:cab}
        \mathcal{S}_{CAB} = softmax(QK^T)V + S_{skip}
\end{equation}

\noexpand where $\mathcal{S}_{skip} = Q = K = V = Flatten(\mathcal{S})$ and $Flatten(\cdot)$ is the module that flattens features on the feature size axis. A skip connection is added to Equation~\eqref{equa:cab} to mitigate the vanishing gradient issue and facilitate effective learning by reutilizing the previous features. Similarly, a Spatial Attention Block (SAB) can be achieved through self-attention by treating the channel as a token dimension and the feature size as the number of tokens. However, this approach results in an attention matrix $\mathbb{R}^{HW\times HW}$, which then requires multiplication by the value matrix to compute the attention. This process can be computationally expensive and slow, making it less suitable for high-resolution images and real-world applications. Fortunately, this issue can be mitigated by incorporating SAM~\cite{sanghyun2018cbam}, which is much faster than SAB and does not depend on the feature size. The idea of SAM is depicted in Fig.~\ref{fig:mask_decoder}d, which can be expressed as follows:
\begin{equation}
    \mathcal{S}_{SAM} = \sigma\left(f^{3\times3}([AvgPool(\mathcal{S});MaxPool(\mathcal{S})])\right) \times \mathcal{S}_{skip}
\end{equation}

where $\sigma$ is sigmoid activation, $f^{3\times3}$ represents a convolution operation with filter size of $3\times3$. By leveraging SAM for channel attention, DMA can mitigate the weak long-range dependencies in CNN architectures at the feature level.

\subsubsection{Feature and Token attention}
Another key feature of DMA is its capability to align the query tokens with the image features using cross-attention~\cite{chunfu2021cross_attention}. We propose the utilization of cross-attention for each image feature in query tokens, namely CTFA, enabling the integration of current image features into the query tokens. CTFA employs attention to integrate information from image features to query tokens, where the query tokens are considered as $Q\in \mathbb{R}^{F\times N}$, and the image features from CAB or SAM is represented as $K\in \mathbb{R}^{F\times HW}$ and $V\in \mathbb{R}^{F\times HW}$ in formula~\eqref{equa:ctfa}.
\begin{equation}
    \label{equa:ctfa}
    QT_{attention} = softmax\left(\frac{Q' K'^T}{\sqrt{d_k}}\right)V'
\end{equation}

where $Q'$ = $Q^TW^q$, $K'$ = $K^TW^k$, $V'$ = $V^TW^v$ are the projection of $Q,K,V$ with learnable matrix $W^q, W^k, W^v$ and $d_k$ is the feature dimension of $K'$. The final query tokens, which encompass feature attention information, are obtained by summing the query tokens and the outputs of the two CTFA modules applied to CAB and SAM as follows:
\begin{equation}
    \label{equa:query_token}
    QT = CTFA\left(QT, \mathcal{S}_{CAB}\right) + CTFA\left(QT, \mathcal{S}_{SAM}\right) + QT_{skip}
\end{equation}

After improving the representation of feature embedding and query tokens through DMA, we apply the Cross-Feature Token Attention (CFTA), which is similar to CTFA, to apply token attention to features. In CTFA, the query tokens will be considered as $K\in \mathbb{R}^{F\times N}$ and $V\in \mathbb{R}^{F\times N}$, while the feature embeddings are represented as $Q\in \mathbb{R}^{F\times HW} = \mathcal{S}_{CAB} + \mathcal{S}_{SAM}$. The process from Equation~\eqref{equa:cab} to~\eqref{equa:query_token} is iterated with $h$ blocks before proceeding to another CTFA for query token alignment. Finally, the output of the DMAD will be fed into the upsampling and MLP blocks to project the feature dimensions to the previous stage.

\subsection{Multi-Level Feature Distribution}
In general, the current architecture of QTSeg demonstrates high performance with low computational complexity. However, its performance still lags behind recent approaches that need further improvement. As shown in Table~\ref{tab:ablation}, employing a single decoder head does not yield performance as high as when utilizing multiple decoder heads. This highlights the significant contributions of features extracted from different stages to the query mask decoder. We hypothesize that the overall prediction quality can be enhanced by ensuring each query mask decoder receives a sufficient amount of valuable features for mask prediction computation. Based on this assumption, we designed an MLFD approach, which combines features containing information from all encoder stages and produces new features at each stage.

The MLFD module functions by distributing features through a process that involves downsampling high-level features to low-level ones and then concatenating them with the existing low-level features. Additionally, the low-level features are upsampled and concatenated with the high-level features. The downsampling block ($DOWN$) consists of a 2D convolutional layer, batch normalization, and SiLU activation layer, while the upsampling block ($UP$) follows a similar design but replaces the 2D convolutional layer with a 2D convolutional transpose layer. A 1x1 2D convolutional layer ($PROJ$) is also used to reduce the current feature size by half to maintain the current feature size after concatenating with other features. In each stage, the current stage comprises a larger proportion of features (larger channel dimension) than the other stages, with 50\% of feature size $F$ for the current stage and 25\% of feature size $F$ for the remaining stages. This distribution is selected to ensure that all features, which have feature size divisible by 4, can be effectively divided into three outputs of the encoder. As a result, only a 1:1:2 ratio is suitable for dividing and concatenating the encoder's output. Subsequently, the outputs of the MLFD can be obtained in the following manner:
\begin{equation}
    \mathcal{SF}_0 = Concat(PROJ(\mathcal{S}_0), UP(\mathcal{S}_1), UP(\mathcal{S}_2))
\end{equation}
\begin{equation}
    \mathcal{SF}_1 = Concat(DOWN(\mathcal{S}_0), PROJ(\mathcal{S}_1), UP(\mathcal{S}_2))
\end{equation}
\begin{equation}
    \mathcal{SF}_2 = Concat(DOWN(\mathcal{S}_0), DOWN(\mathcal{S}_1), PROJ(\mathcal{S}_2))
\end{equation}
\noexpand where $\mathcal{SF}_i$ is the output fusion features at stage $i^{th}$ which have the same dimension as $\mathcal{S}_i$. The inclusion of MLFD to reorganize the current features results in enhanced overall performance with minimal computational demand, as shown in Setting~6 of Table~\ref{tab:ablation}.

\section{Experimental Results and Discussion}

\label{sec:experiments}
\subsection{Datasets}
\subsubsection{Lesion Segmentation}
The ISIC2016~\cite{isic2016} dataset specifically focuses on enhancing melanoma diagnosis through the utilization of high-quality, human-validated datasets comprising skin lesion images. The challenge provided official training and testing datasets featuring 900 dermoscopic lesion images for training and 379 for testing. These images were accompanied by ground truth masks labeled by experts and saved in binary mask format. The BKAI-IGH NeoPolyp~\cite{neopoly} is the polyp segmentation dataset released by the BKAI Research Center at Hanoi University of Science and Technology in collaboration with the Institute of Gastroenterology and Hepatology (IGH) in Vietnam. This dataset comprises 1,200 endoscopic images, with a division of 1,000 images for training and 200 images for testing purposes. While an official split of 1,000 training and 200 testing samples exists, the official test ground truth is not public due to the ongoing challenge. Thus, we partition the training set into five folds to evaluate the model's performance comprehensively. To ensure the reproducibility of results and facilitate future comparisons, we first sort all samples based on their sample names. Subsequently, these samples were segmented into K folds by array slicing. In our experimental setup, we executed the models across five folds and subsequently reported the averaged metrics for a comprehensive evaluation.

\subsubsection{Breast Cancer Segmentation}
The BUSI~\cite{busi} dataset comprises a collection of breast ultrasound images obtained from 600 female patients aged between 25 and 75 years. This dataset includes 780 images categorized into three classes: normal, benign, and malignant. As the normal images do not exhibit any lesions, a new dataset was obtained by excluding these normal images. The refined dataset exclusively contains cases classified as benign (437 images) and malignant (210 images), which were then partitioned using the K-fold strategy. 

\subsubsection{Cell Segmentation}
The DSB2018~\cite{caicedo2019nucleus} was published in 2018, focusing on nucleus segmentation across many imaging experiments. This diverse collection includes images under various conditions, showcasing differences in cell type, magnification, and imaging modalities such as brightfield and fluorescence. Each image is annotated with a unique identifier, linking it to corresponding segmentations, which are provided in training datasets. We also partition the training set into five folds to evaluate the model's performance comprehensively.

\subsubsection{Retinal Vessel Segmentation}
The FIVES~\cite{jin2022fives} aimed to develop and assess artificial intelligence models for segmenting blood vessels in fundus images. This dataset comprises a total of 800 samples with high-quality color and resolution. The primary goal is to enable precise, efficient, and automated segmentation of retinal vessels, a crucial step in diagnosing and managing a range of clinical conditions. The official training and testing datasets include 600 fundus images for training and 200 fundus images for testing.

\subsection{Implementation Details and Evaluation Metrics}
\subsubsection{Implementation Details}
To mitigate model overfitting during training, we implemented the augmentation process described in MISSFormer~\cite{missformer}. Our QTSeg model underwent training for 350 epochs on each dataset, using a batch size of 32 and the AdamW optimizer with a learning rate set to 0.001 and a weight decay of 0.0001. A scheduler was employed to adjust the learning rate, reducing it by a factor of 0.5 every 50 epochs. For retinal vessel segmentation specifically, a batch size of 4, the SGD optimizer with an initial learning rate of 0.01, and a poly scheduler strategy were employed. Input images were resized to 512 x 512 for lesion, breast cancer, and cell segmentation and 1024 x 1024 for retinal vessel segmentation. Following resizing, these images were normalized to a value range of [0,1] using min-max normalization. To maintain fairness in comparisons, no post-processing was applied to the output results. All experiments were conducted on an NVIDIA GeForce RTX 3090 GPU running on the Debian 12 system. For evaluation metrics, we utilized standard metrics, including Mean Absolute Error (MAE), Accuracy (Acc), Intersection-over-Union (IoU), and Dice Similarity Coefficient (Dice).

\subsection{Comparisons with Other Methods}
\label{exp:results}

\begin{table*}[ht!]
\centering
\caption{Performance comparison between QTSeg and other state-of-the-art methods. The bold font signifies the \textbf{best result}, while the underlined font denotes the \underline{second-best result}.}
\label{tab:lesion_performance}
\setlength{\tabcolsep}{3pt}
\begin{tabular}{l|rr|ccc|ccc|ccc|ccc}
\hline
           
 \multirow{2}{*}{\textbf{Method}} &\multirow{2}{*}{\textbf{Prams}}  &\multirow{2}{*}{\textbf{FLOPs}} & \multicolumn{3}{c}{\textbf{ISIC2016}} & \multicolumn{3}{|c|}{\textbf{BUSI}} & \multicolumn{3}{c|}{\textbf{BKAI-IGH NeoPolyp}} & \multicolumn{3}{c}{\textbf{DSB2018}}\\
 \cline{4-15}
                                       &&&\textbf{MAE}$\downarrow$ & \textbf{Dice}$\uparrow$ & \textbf{IoU}$\uparrow$& \textbf{MAE}$\downarrow$ & \textbf{Dice}$\uparrow$ & \textbf{IoU}$\uparrow$& \textbf{MAE}$\downarrow$ &\textbf{Dice}$\uparrow$ & \textbf{IoU}$\uparrow$& \textbf{MAE}$\downarrow$ &\textbf{Dice}$\uparrow$ & \textbf{IoU}$\uparrow$\\ 
\hline                                              % ISIC2016              % BUSI                  % BKAI                  % DSB2018     
U-Net (2015)~\cite{unet}           &23.63 M&33.39 G &4.98 &89.56 & 83.61  & 4.09 & 77.17 & 68.47  & 0.99 & 87.25 & 81.33  & 3.12 & 88.83 & 81.04   \\ 
UNet++ (2020)~\cite{unet++}        &24.38 M&35.60 G &5.12 &89.46 & 83.63  & \underline{3.53} & 78.58 & 69.65  & 0.92 & 88.44 & 82.89  & 3.00 & 89.22 & 81.63   \\
EGE-UNet (2023)~\cite{egeunet}     &\textbf{0.05 M} &\textbf{0.33 G}  &5.68 &88.98 & 81.74  & 4.85 & 68.24 & 57.11  & 3.20 & 59.98 & 49.28  & 3.21 & 85.29 & 76.03   \\ 
MISSFormer (2023)~\cite{missformer}&42.33 M&109.45 G&5.45 &88.37 & 80.75  & 4.25 & 78.63 & 69.66  & 0.95 & 87.98 & 81.89  & 2.28 & 89.78 & 82.36   \\
Swin-Unet (2023)~\cite{swinunet}   &27.27 M&36.98 G &4.74 &90.12 & 83.21  & 3.80 & 81.35 & 72.90  & 0.73 & 90.80 & 85.24  & 2.27 & 89.95 & 82.67   \\   
MALUNet (2023)~\cite{malunet}      &\underline{0.18 M} &\underline{0.37 G}  &4.55 &92.01 & 85.20  & 4.72 & 75.14 & 60.26  & 1.84 & 82.09 & 69.67  & 5.56 & 81.72 & 71.96   \\
H2Former (2023)~\cite{h2former}    &33.71 M&33.56 G &\underline{3.80} &\underline{92.41} & 86.45  & 4.14 & 74.06 & 63.64  & 1.69 & 73.83 & 64.29  & 3.00 & 89.17 & 81.33   \\ 
MHorUNet (2024)~\cite{mhorunet}    &4.96 M &2.38 G  &4.74 &91.13 & 84.91  & 4.11 & 77.24 & 67.89  & 2.12 & 64.66     & 53.90      & 3.34 & 83.14 & 73.28   \\
VM-UNetV2 (2024)~\cite{vmunetv2}   &22.77 M&4.40 G  &4.43 &92.14 & 85.43  & 5.87 & 67.80 & 51.63  & 1.26 & 87.71 & 78.16  & 2.56 & 88.46 & 80.44   \\
MET-Net (2024)~\cite{metnet}       &18.65 M&15.60 G &4.06 &91.67 & 85.66  & 3.98 & 78.54 & 69.80  & 0.74 & 91.05 & 85.99  & \underline{2.18} & \underline{90.30} & \underline{83.10}   \\
TransUNet (2024)~\cite{transunet}  &109.54 &56.66 G &4.23 &91.31 & 84.96  & 3.64 & 81.25 & \underline{73.10}  & \underline{0.69} & \underline{91.37} & \underline{86.25}  & 2.80 & 89.97 & 82.51   \\
FSCA-Net (2024)~\cite{fscanet}     &43.50 M&32.95 G &5.01 &90.51 & 84.09  & 3.61 & 81.25 & 72.95  & 0.89 & 89.45 & 83.99  & 2.24 & 89.79 & 82.67   \\
% MTUnet &&&&&\\
MedSAM (2024)~\cite{medsam}        &9.79 M &39.91 G &3.88 &92.33 & \underline{86.59} & 3.72 & \underline{81.65} & 72.88 & 0.81 & 88.92 & 83.55 & 2.58 & 88.08 & 80.36\\
\hline
\textbf{QTSeg (Ours)}              &9.21 M &2.60 G  &\textbf{3.00}& \textbf{92.45}& \textbf{86.92}  & \textbf{3.08} & \textbf{81.82} & \textbf{73.46}  & \textbf{0.50} & \textbf{92.71} & \textbf{88.41}  & \textbf{1.94} & \textbf{91.56} & \textbf{85.07}   \\
\hline
\end{tabular}
\end{table*}

\begin{figure*}[!t]
\setlength{\tabcolsep}{2pt}
\centering
\fontsize{9pt}{9pt}\selectfont
\begin{tabular}{cccccccc}

\includegraphics[height=0.115\linewidth, width=0.115\linewidth]{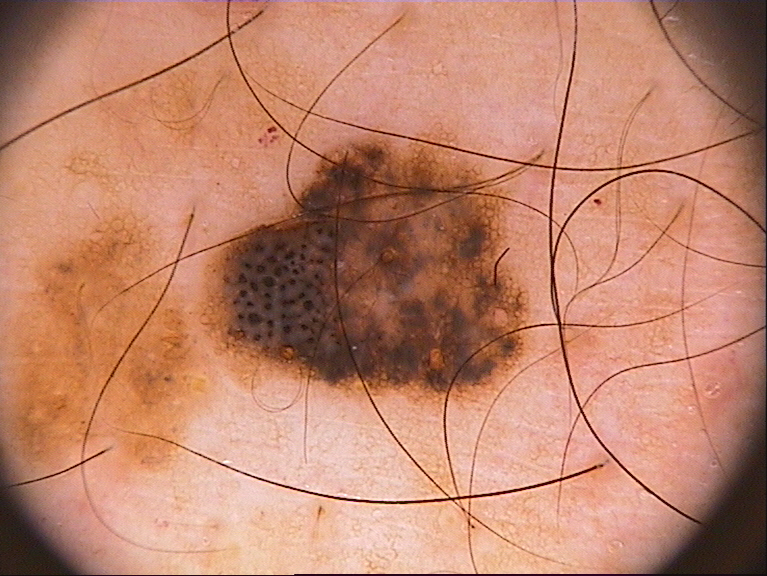} &
\includegraphics[height=0.115\linewidth, width=0.115\linewidth]{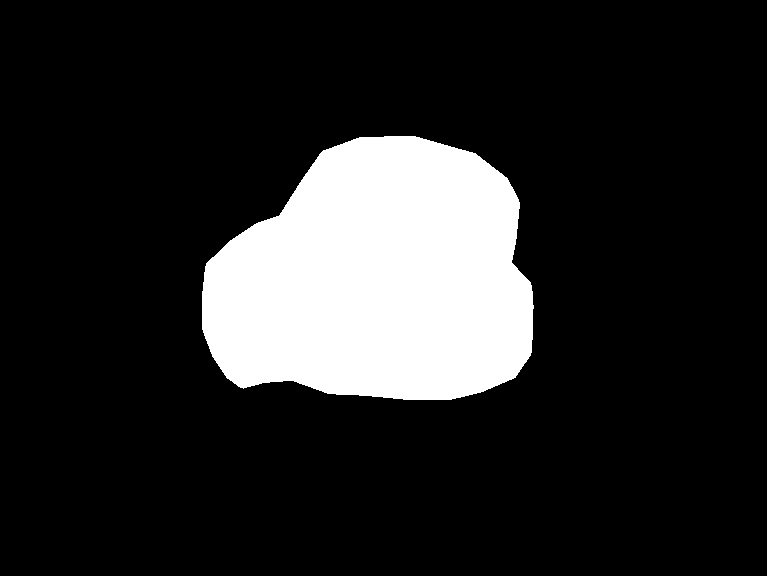} &
\includegraphics[height=0.115\linewidth, width=0.115\linewidth]{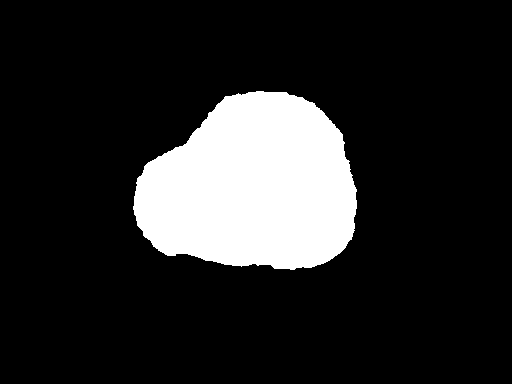} &
\includegraphics[height=0.115\linewidth, width=0.115\linewidth]{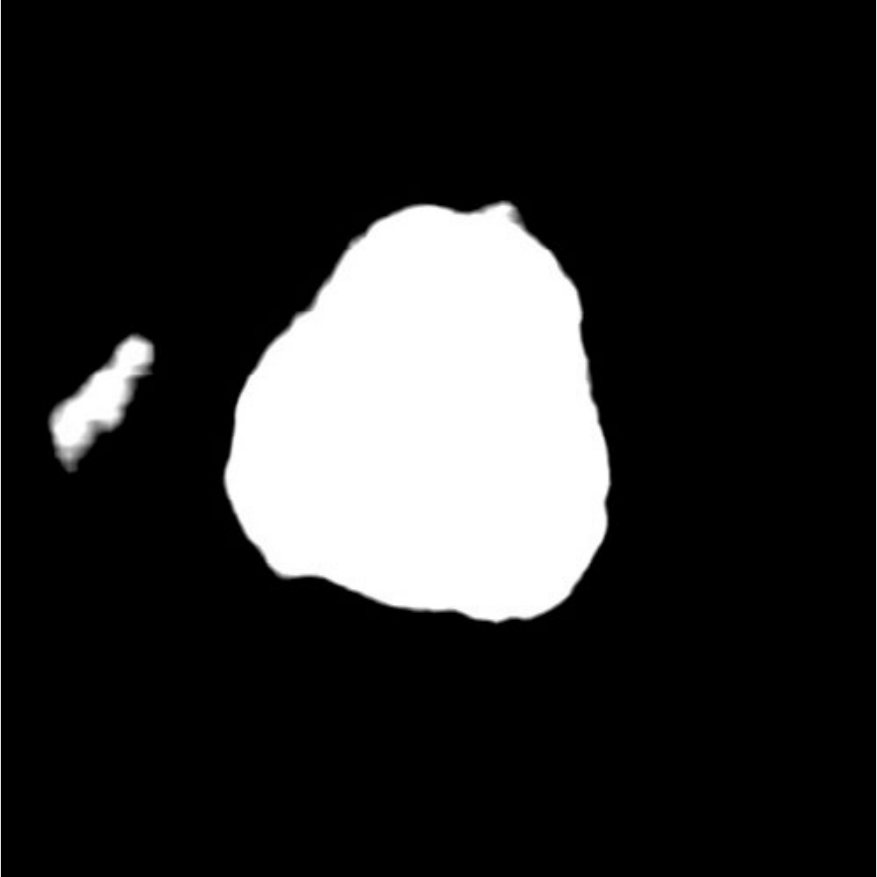} &
\includegraphics[height=0.115\linewidth, width=0.115\linewidth]{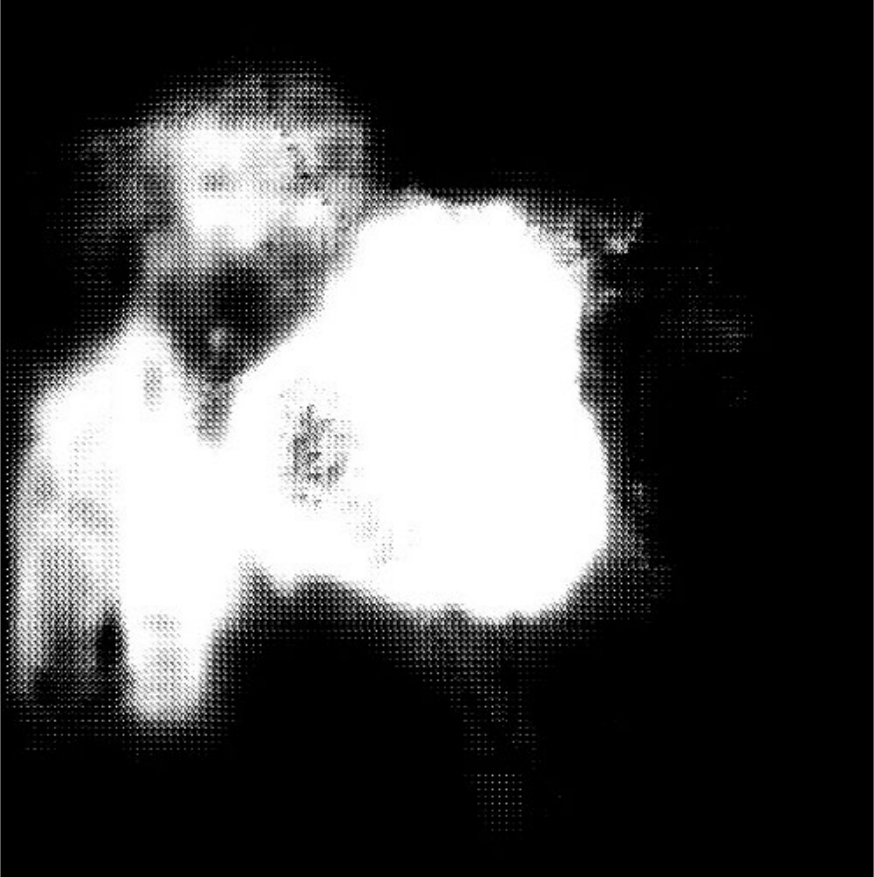} &
\includegraphics[height=0.115\linewidth, width=0.115\linewidth]{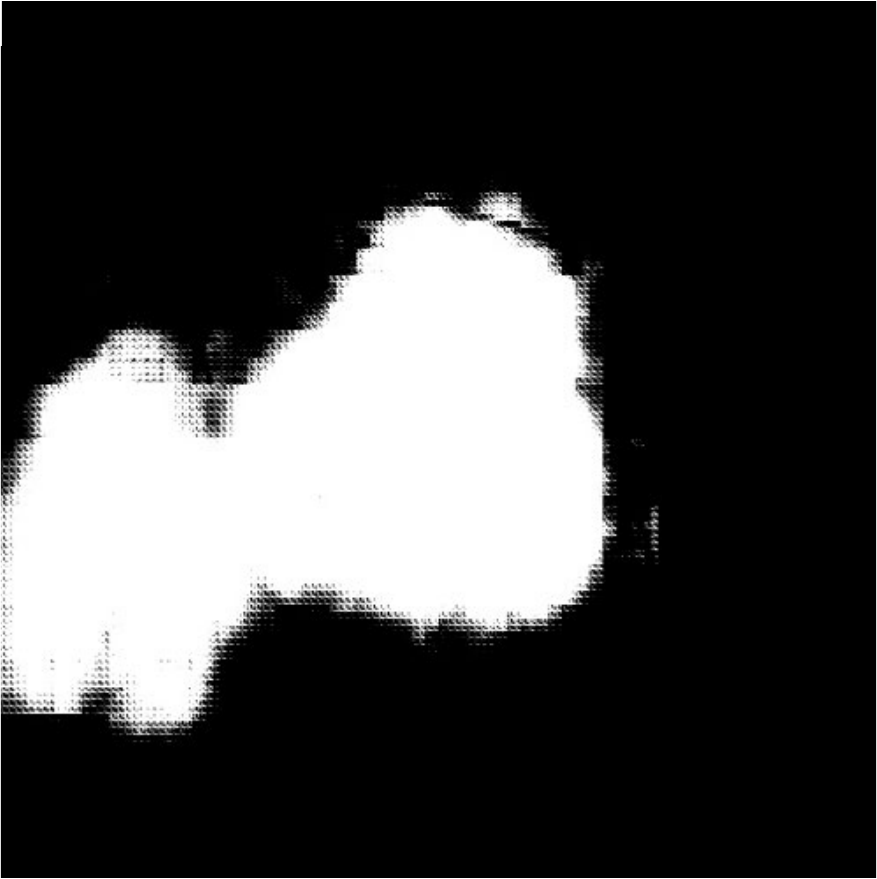} &
\includegraphics[height=0.115\linewidth, width=0.115\linewidth]{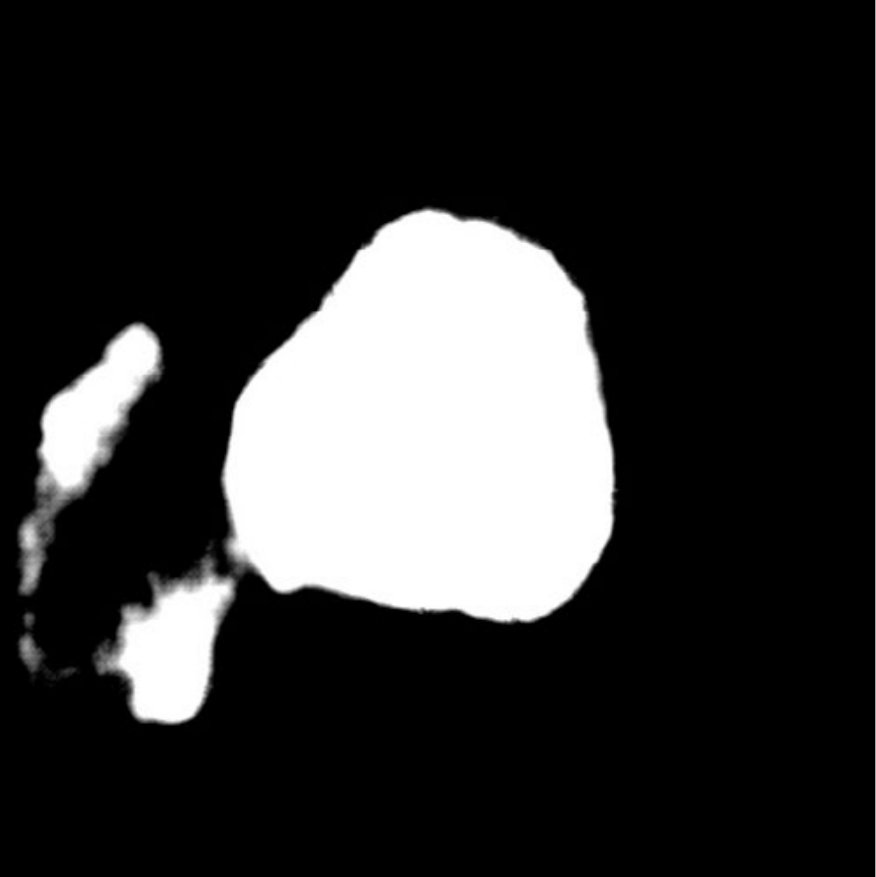} &
\includegraphics[height=0.115\linewidth, width=0.115\linewidth]{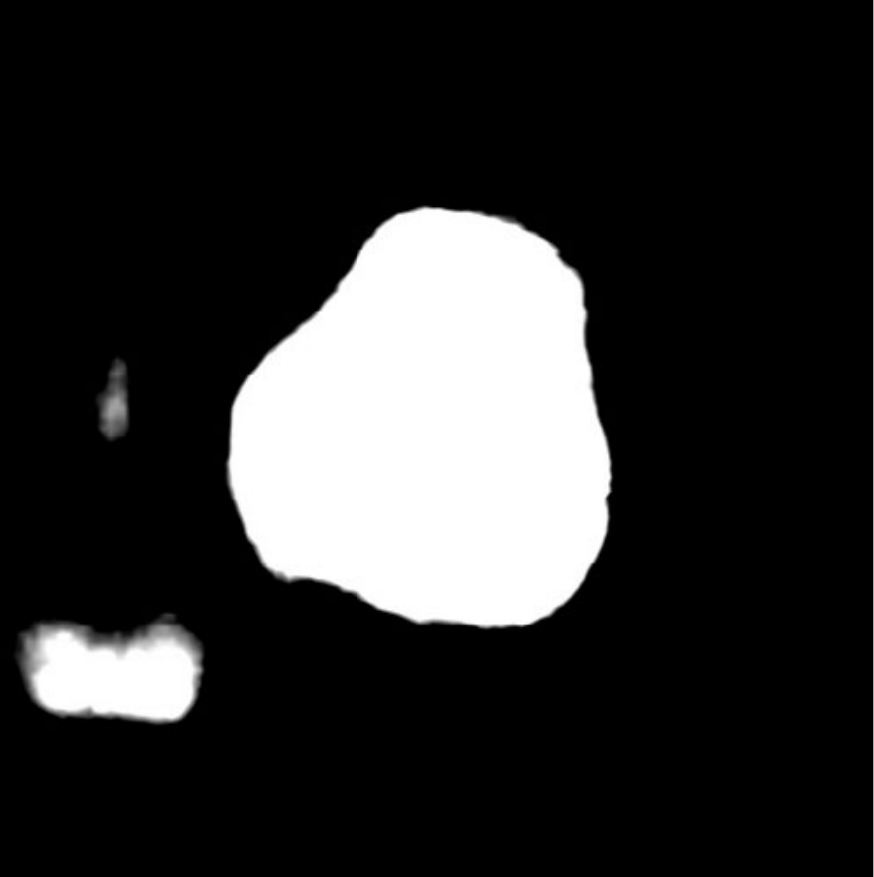} \\

\includegraphics[height=0.115\linewidth, width=0.115\linewidth]{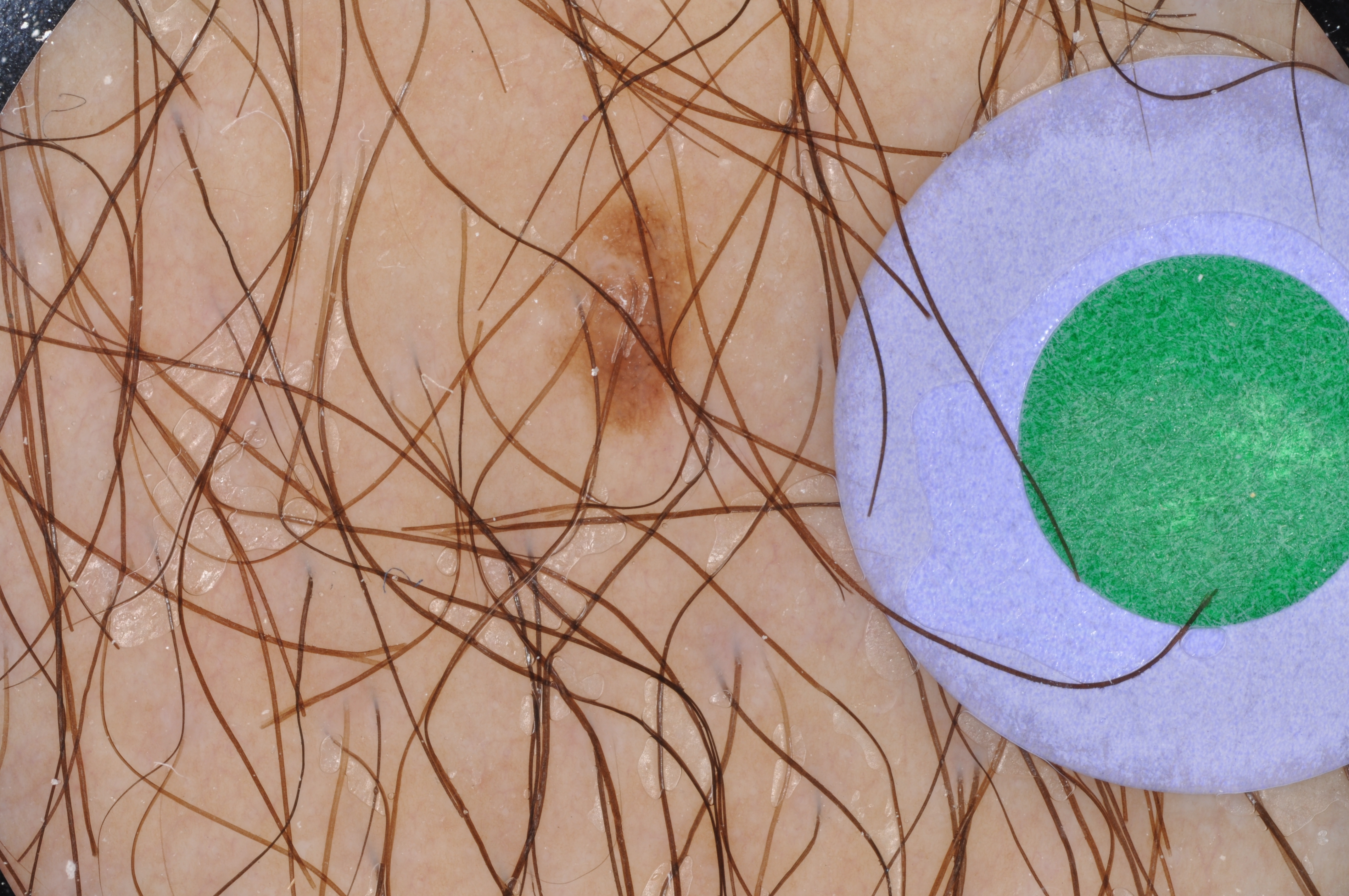} &
\includegraphics[height=0.115\linewidth, width=0.115\linewidth]{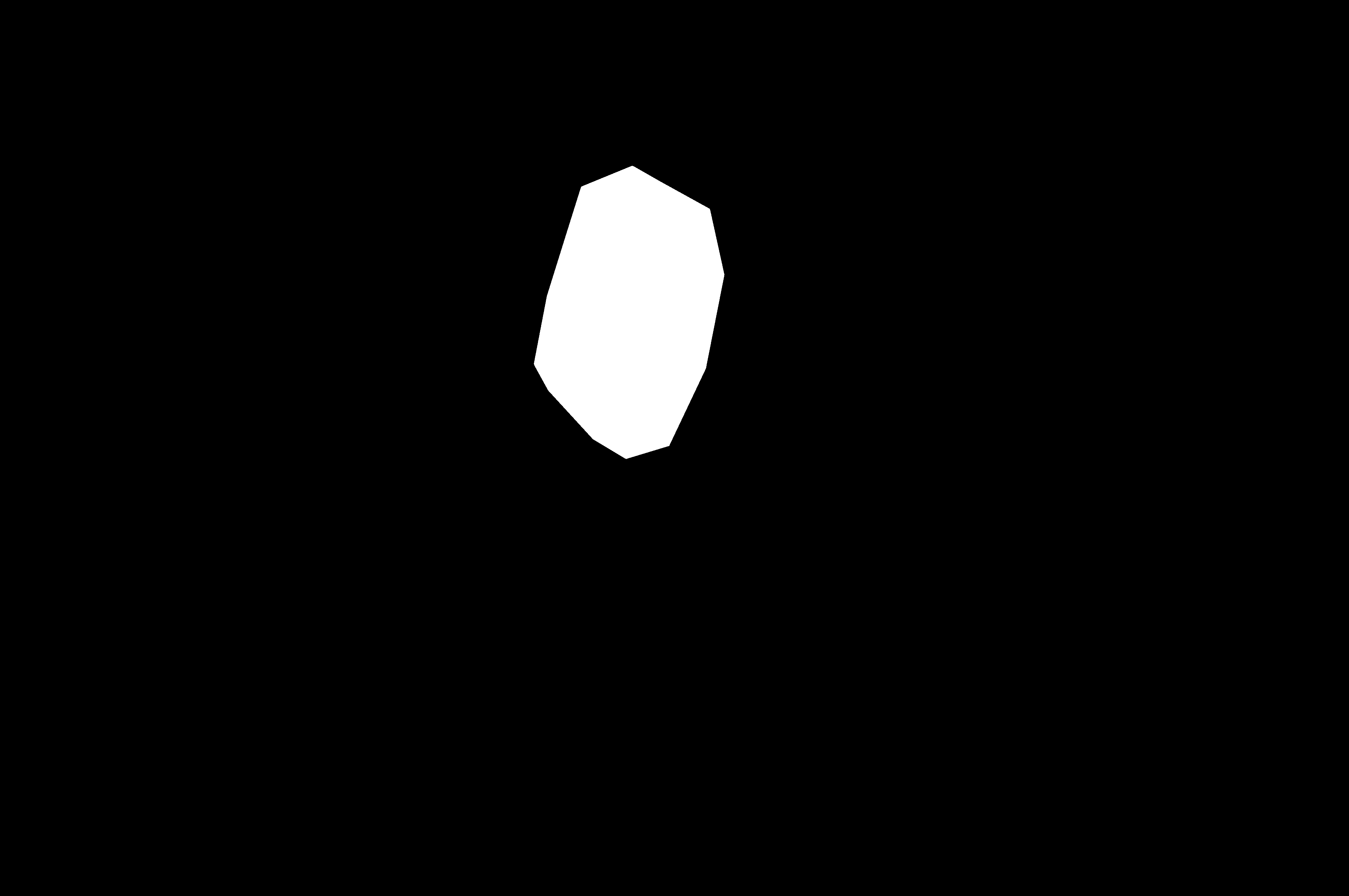} &
\includegraphics[height=0.115\linewidth, width=0.115\linewidth]{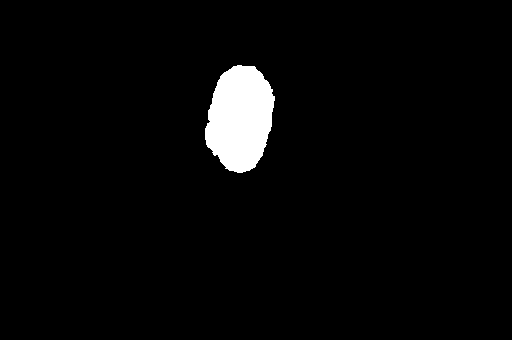} &
\includegraphics[height=0.115\linewidth, width=0.115\linewidth]{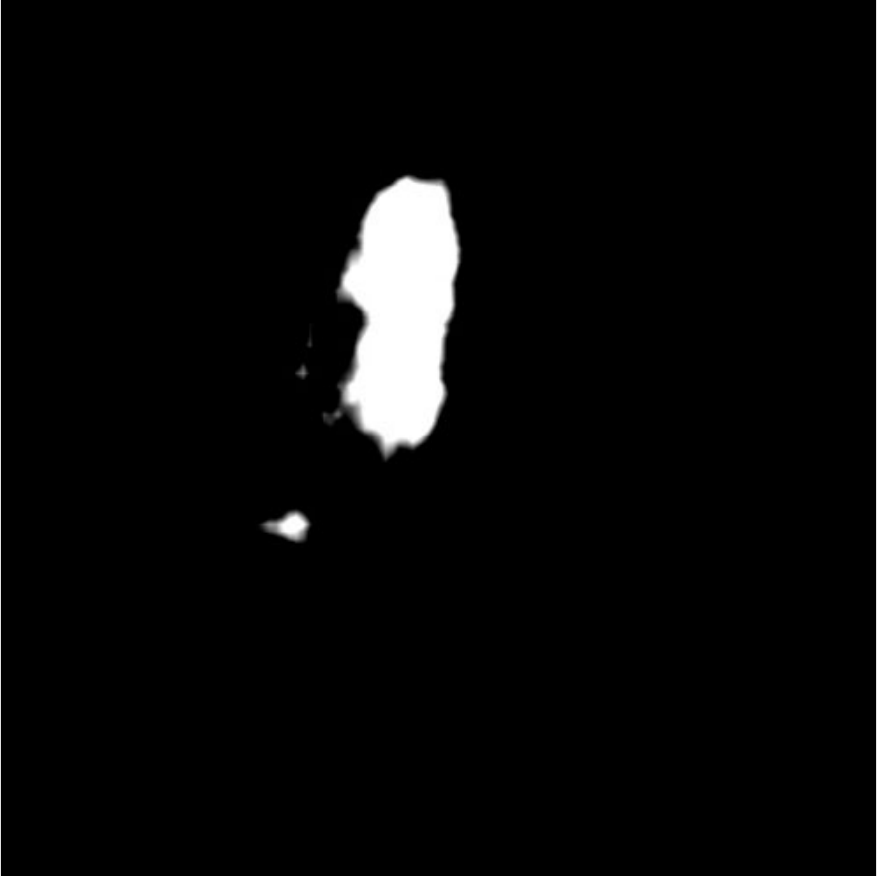} &
\includegraphics[height=0.115\linewidth, width=0.115\linewidth]{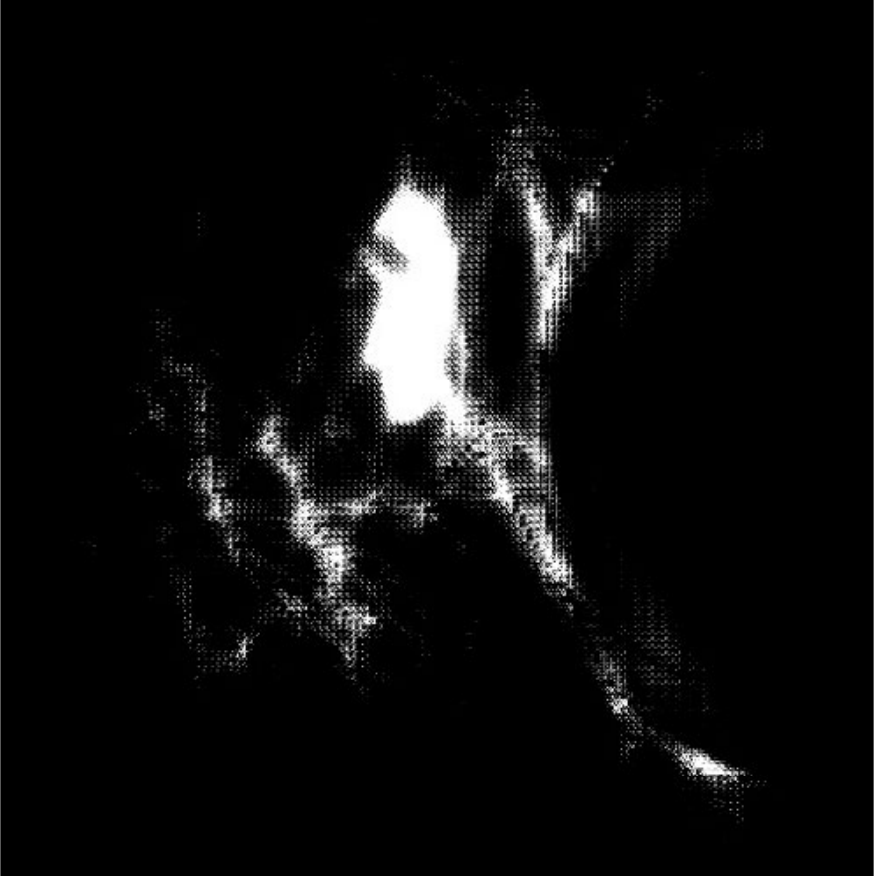} &
\includegraphics[height=0.115\linewidth, width=0.115\linewidth]{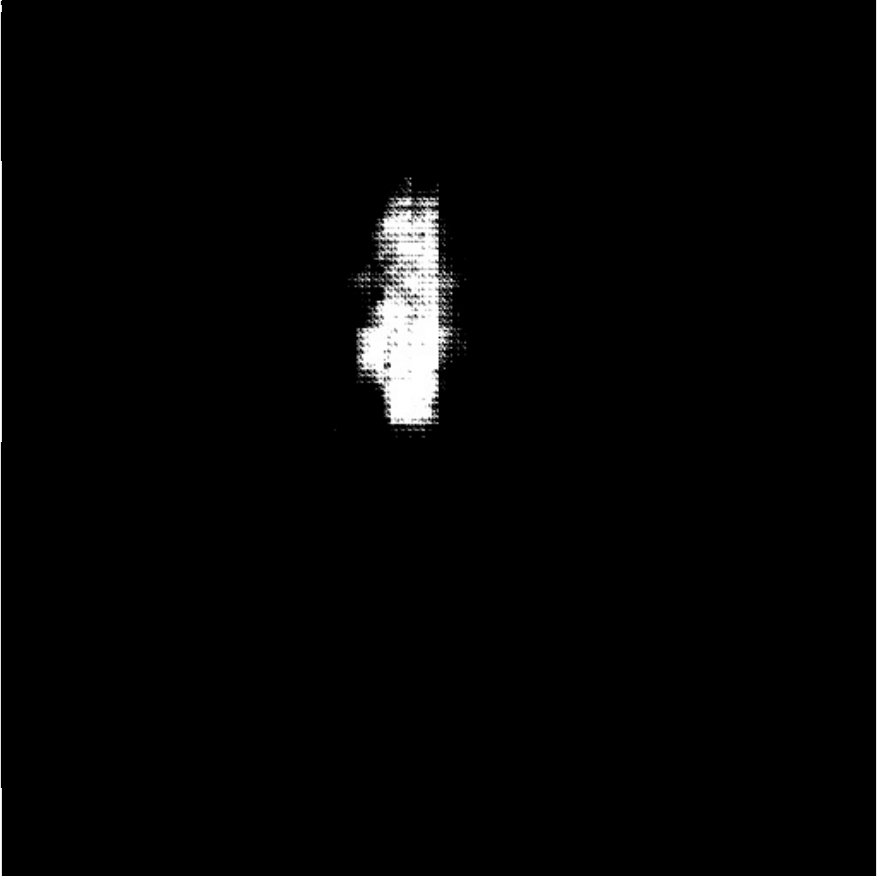} &
\includegraphics[height=0.115\linewidth, width=0.115\linewidth]{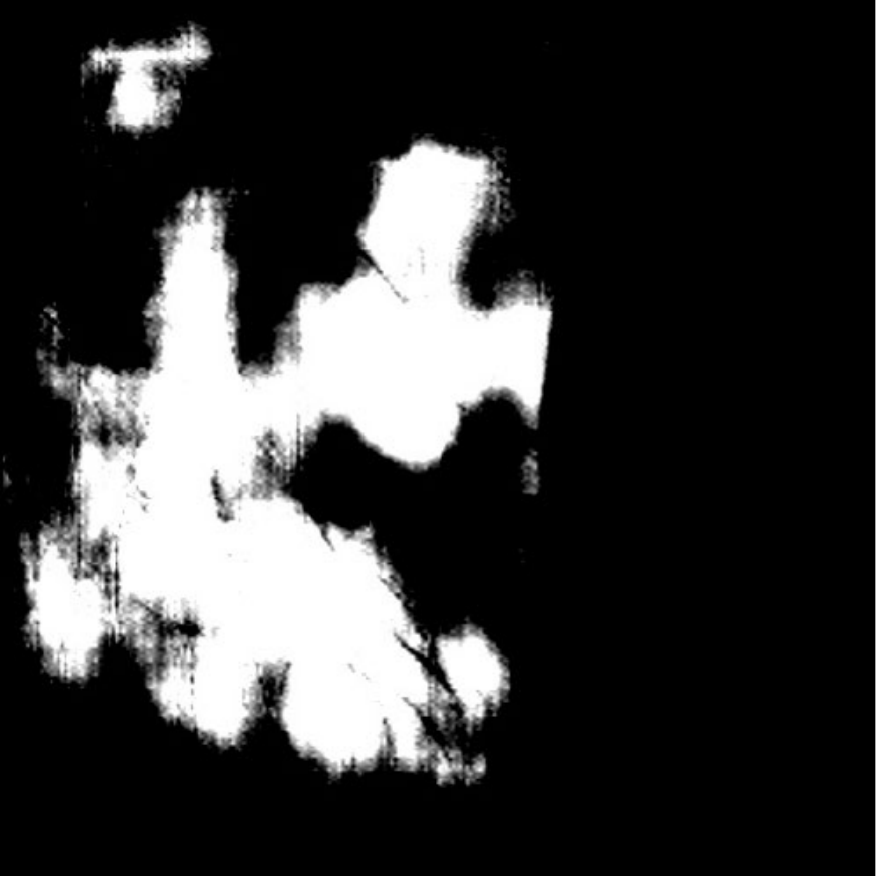} &
\includegraphics[height=0.115\linewidth, width=0.115\linewidth]{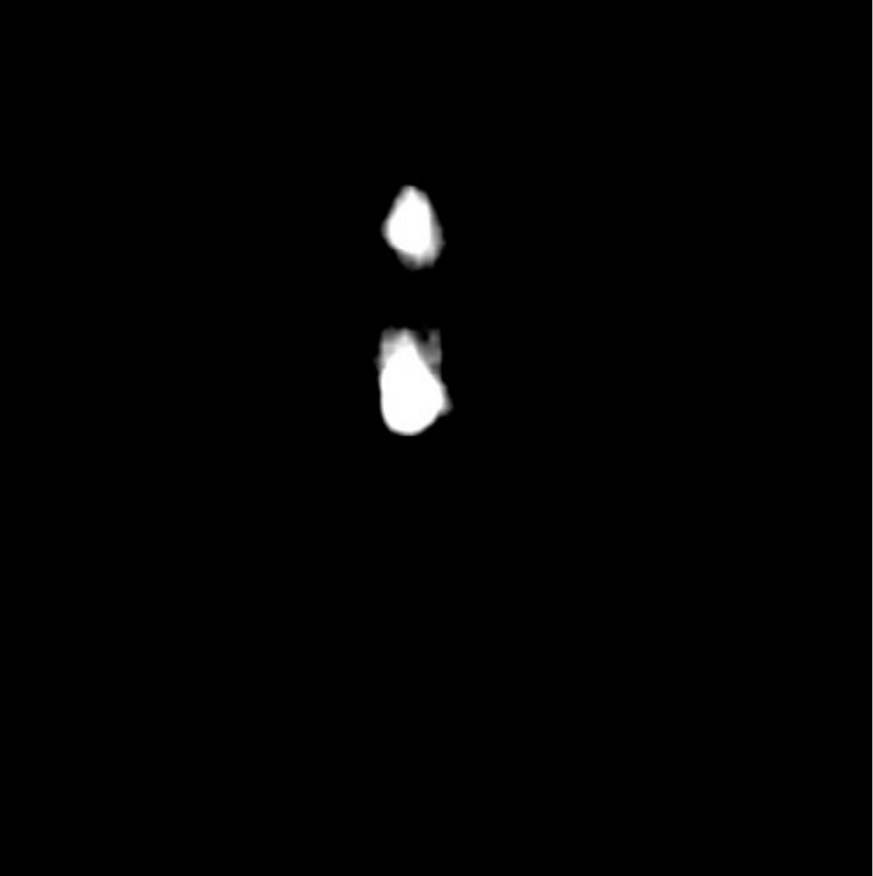} \\ 

(a) Inputs & (b) Ground Truth & (c) \textbf{Ours} & (d) H2Former & (e) MISSFormer & (f) Swin-Unet & (g) TransUnet & (h) DeepLabv3+ \\

\end{tabular}
\centering
\caption{The comparison of visualization prediction between QTSeg and other methods on the ISIC2016 dataset.}
\label{fig:isis_viz}
\end{figure*}

\subsubsection{Results on Lesion Segmentation}

Table~\ref{tab:lesion_performance} presents the performance of QTSeg on the lesion, breast cancer, and cell segmentation datasets, outperforming recent methods across all evaluation metrics. Specifically, our model achieves impressive scores of 92.45\% and 86.92\% in Dice and IoU metrics, respectively, for skin lesion segmentation. For the BKAI-IGH NeoPolyp dataset, our model also attained the highest scores in Dice and IoU metrics, showcasing its effectiveness. Furthermore, %as illustrated in the table, 
QTSeg demonstrates great potential by delivering outstanding results with considerably lower FLOPs and parameter counts compared to other high-accuracy methods. It is worth noting that QTSeg's parameter count is four times lower than that of H2Former~\cite{h2former} while achieving superior performance across all metrics. While EGE-UNet~\cite{egeunet} and MALUNet~\cite{malunet} have smaller parameter sizes, they encounter challenges in converging on the BKAI-IGH NeoPolyp dataset due to inherent design limitations and model parameter constraints. Our approach stands out for achieving precise lesion and polyp segmentation with minimal error compared to alternative methods. 

\begin{figure}
\centering
\fontsize{9pt}{9pt}\selectfont
\begin{tabular}{cccc}

\includegraphics[height=0.23\linewidth, width=0.23\linewidth]{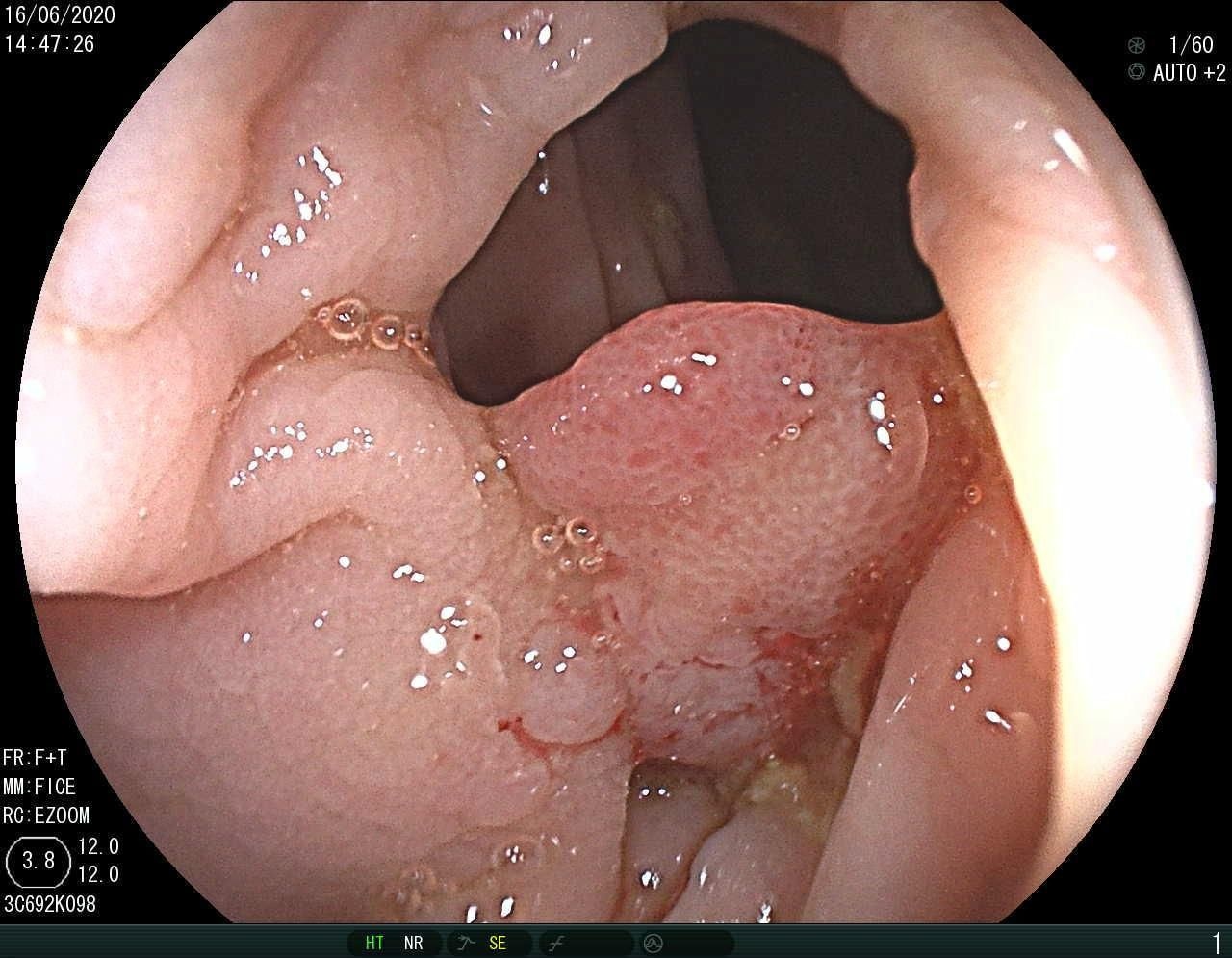} &
\includegraphics[height=0.23\linewidth, width=0.23\linewidth]{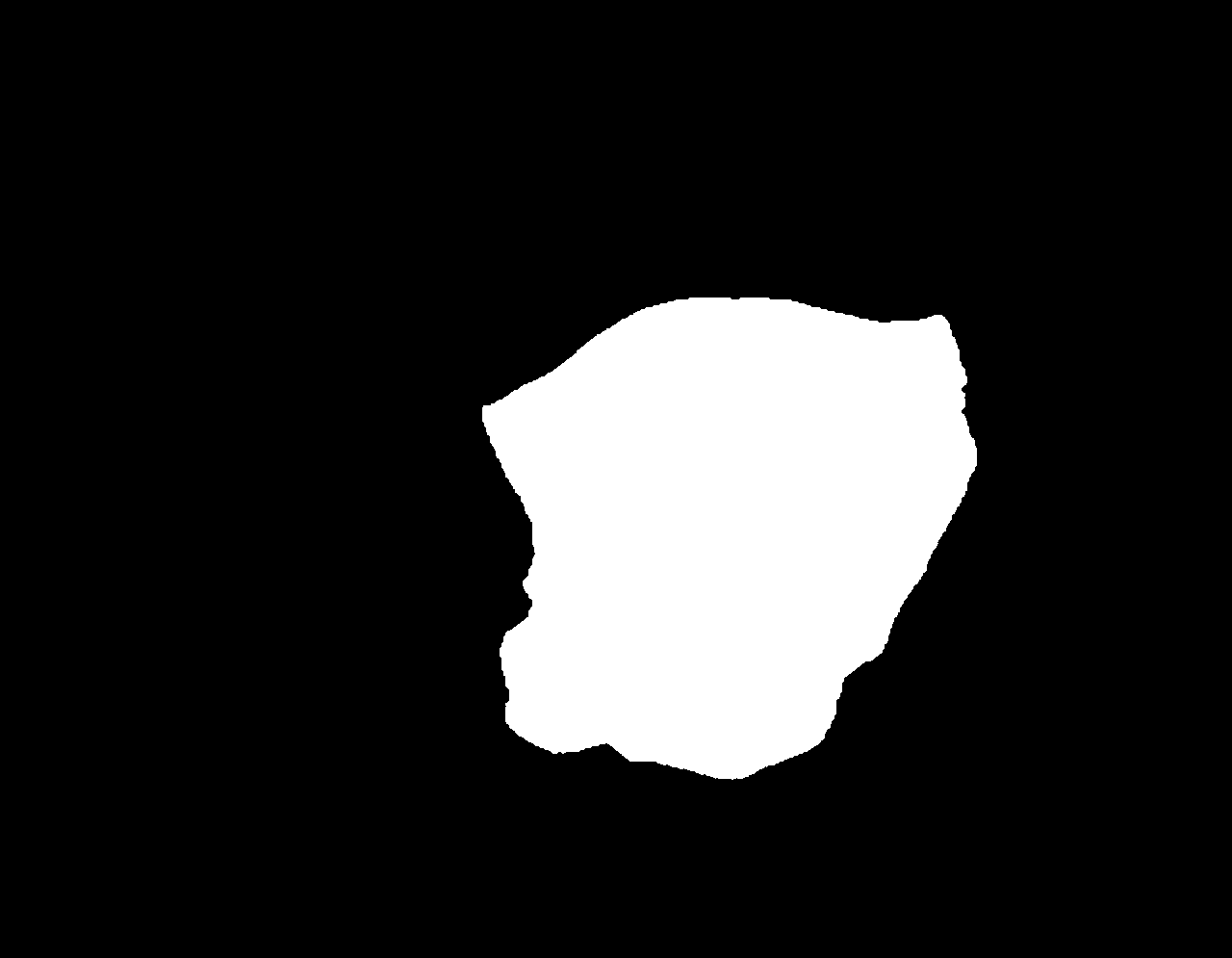} &
\includegraphics[height=0.23\linewidth, width=0.23\linewidth]{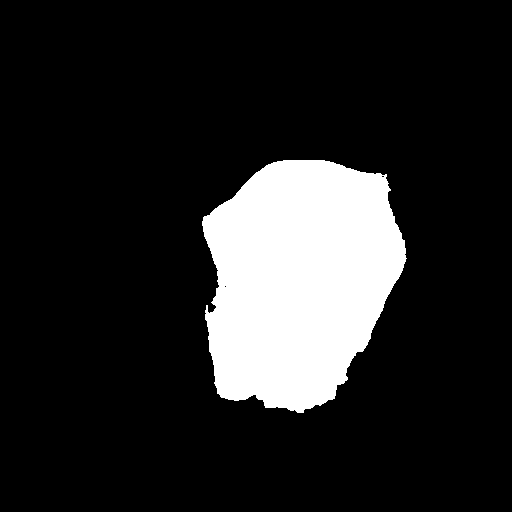} &
\includegraphics[height=0.23\linewidth, width=0.23\linewidth]{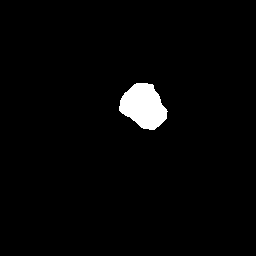} \\

(a) Inputs & (b) Ground Truth & (c) Ours & (d) EGE-UNet \\
\includegraphics[height=0.23\linewidth, width=0.23\linewidth]{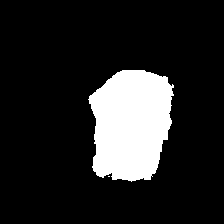} &
\includegraphics[height=0.23\linewidth, width=0.23\linewidth]{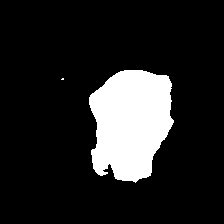} &
\includegraphics[height=0.23\linewidth, width=0.23\linewidth]{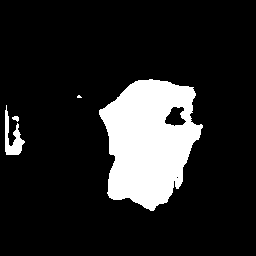} &
\includegraphics[height=0.23\linewidth, width=0.23\linewidth]{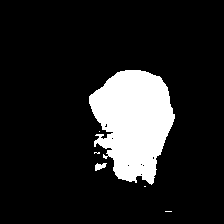} \\

(e) Swin-Unet & (f) MET-Net & (g) VM-UNetV2 & (h) UNet++ \\

\end{tabular}
\caption{The comparison of visualization prediction between QTSeg and other methods on 
% 22af6b2da43f71d4dc3ddad260bfcb44 sample from 
the BKAI-IGH NeoPolyp dataset.
}
\label{fig:bkai_viz}
\end{figure}

On the other hand, as depicted in Fig.~\ref{fig:isis_viz}, the visualization of our model's predictions compared to other methods highlights QTSeg's superior accuracy, with the generated mask closely resembling the ground truth mask. Fig.~\ref{fig:bkai_viz} further demonstrates our model's clean and accurate segmentation mask, emphasizing its effectiveness in medical image segmentation.

\setlength{\tabcolsep}{2pt}

\subsubsection{Results on Cell Segmentation}

The lesion segmentation experiments above mainly consist of single segmentation objects, which do not fully showcase the performance of QTSeg in multiple object segmentation within medical images. To further evaluate the performance of QTSeg in a different segmentation scenario, we conducted an experiment using the DSB2018 dataset. As depicted in Table~\ref{tab:lesion_performance}, QTSeg demonstrates superior performance compared to other approaches, with a significant lead of 1.97\% in the IoU metric over the second-best method. This result underscores the effectiveness of our proposed method in general multiple object segmentation and the specific task of cell segmentation.

\setlength{\tabcolsep}{2pt}
\begin{figure}
\centering
\fontsize{9pt}{9pt}\selectfont
\begin{tabular}{cccc}

\includegraphics[height=0.23\linewidth, width=0.23\linewidth]{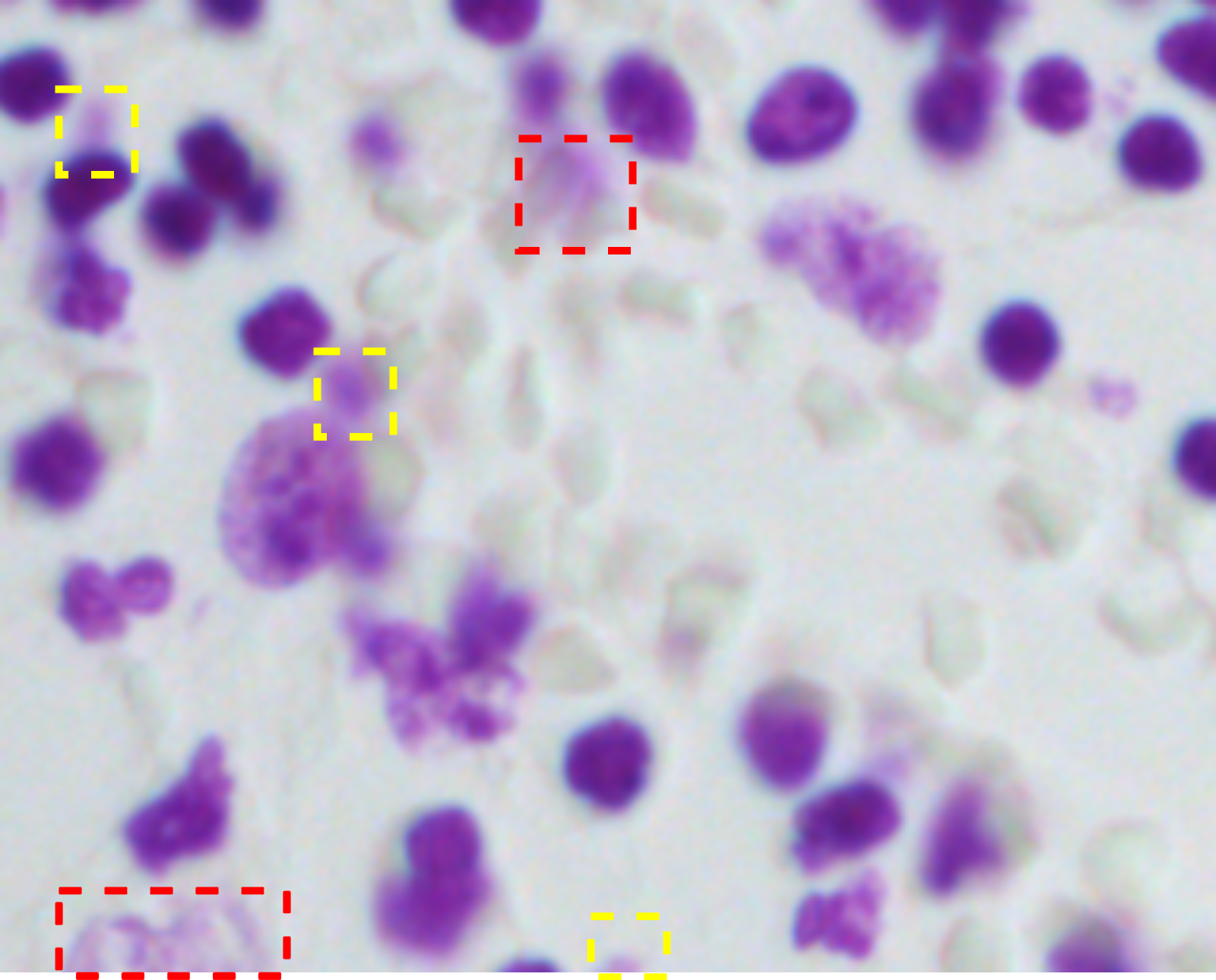} &
\includegraphics[height=0.23\linewidth, width=0.23\linewidth]{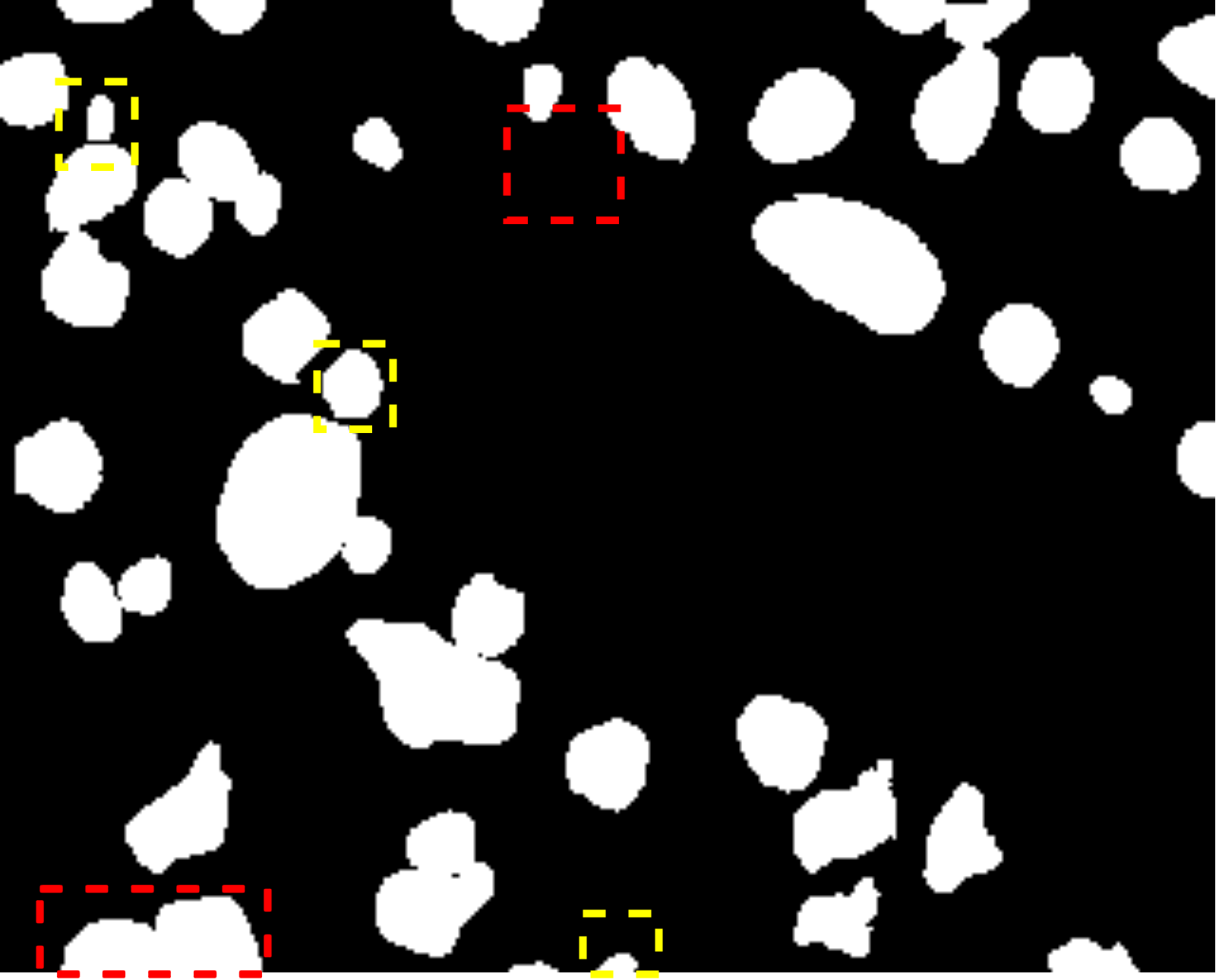} &
\includegraphics[height=0.23\linewidth, width=0.23\linewidth]{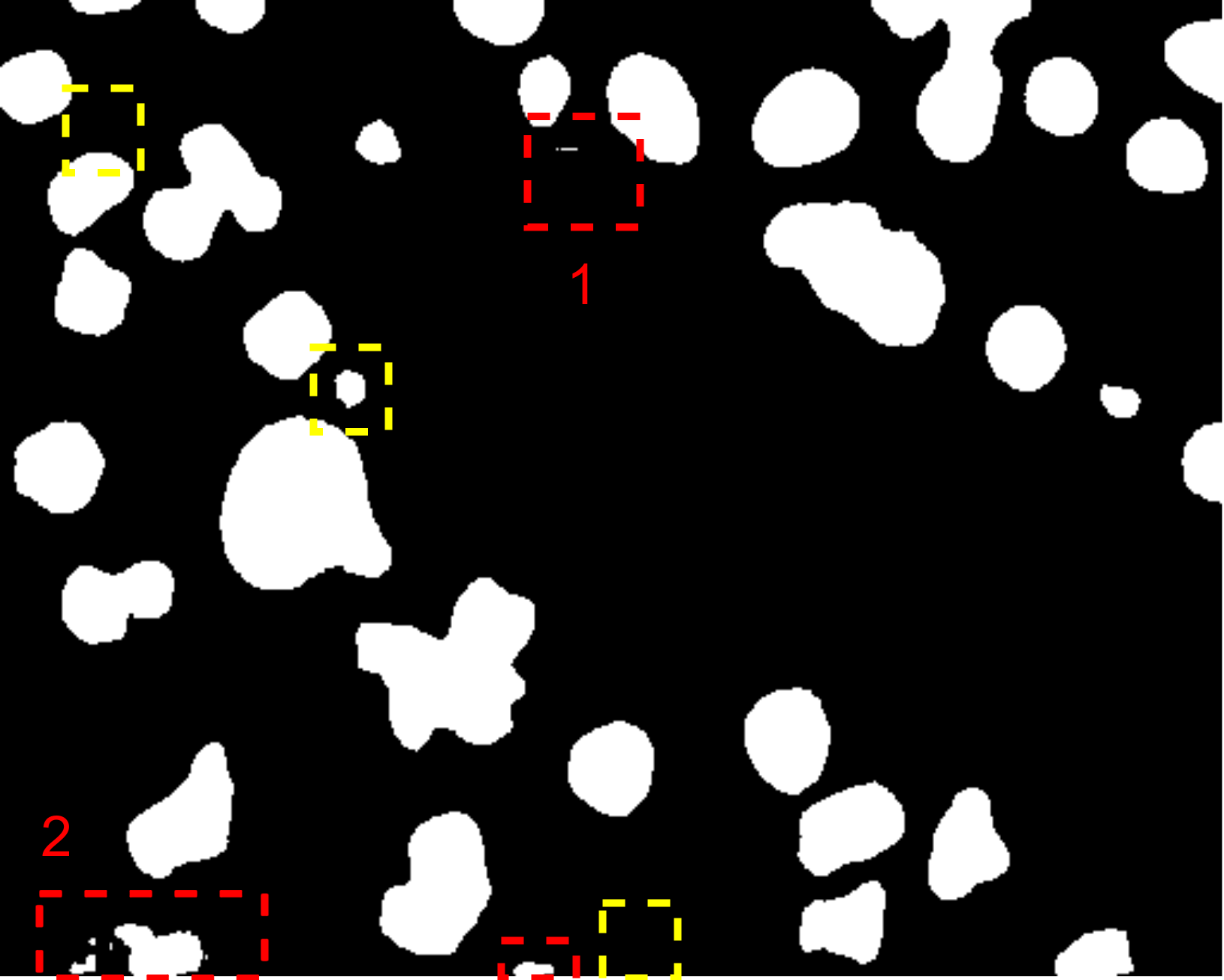} &
\includegraphics[height=0.23\linewidth, width=0.23\linewidth]{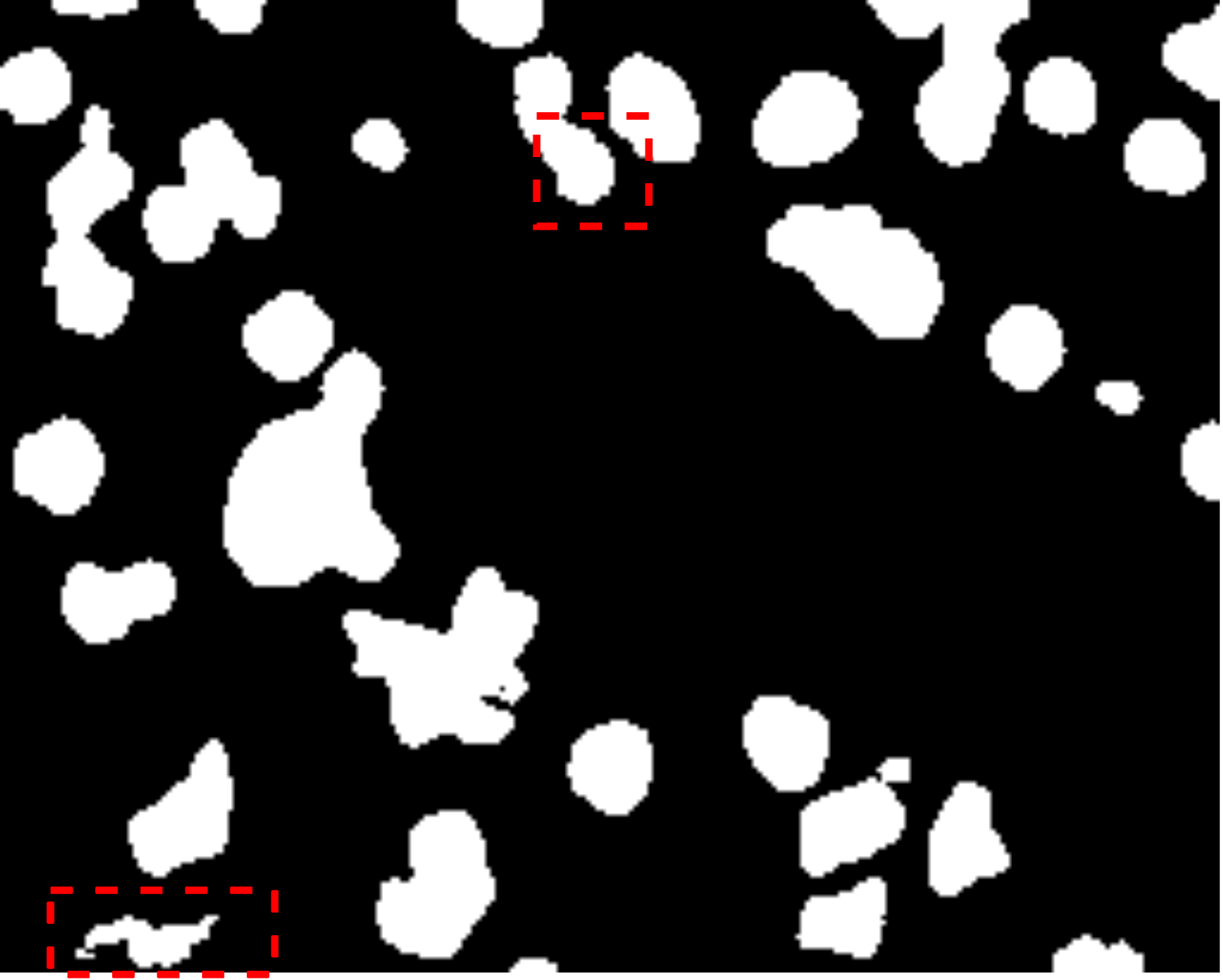} \\

(a) Inputs & (b) Ground Truth & (c) Ours & (d) UNet++ \\
\includegraphics[height=0.23\linewidth, width=0.23\linewidth]{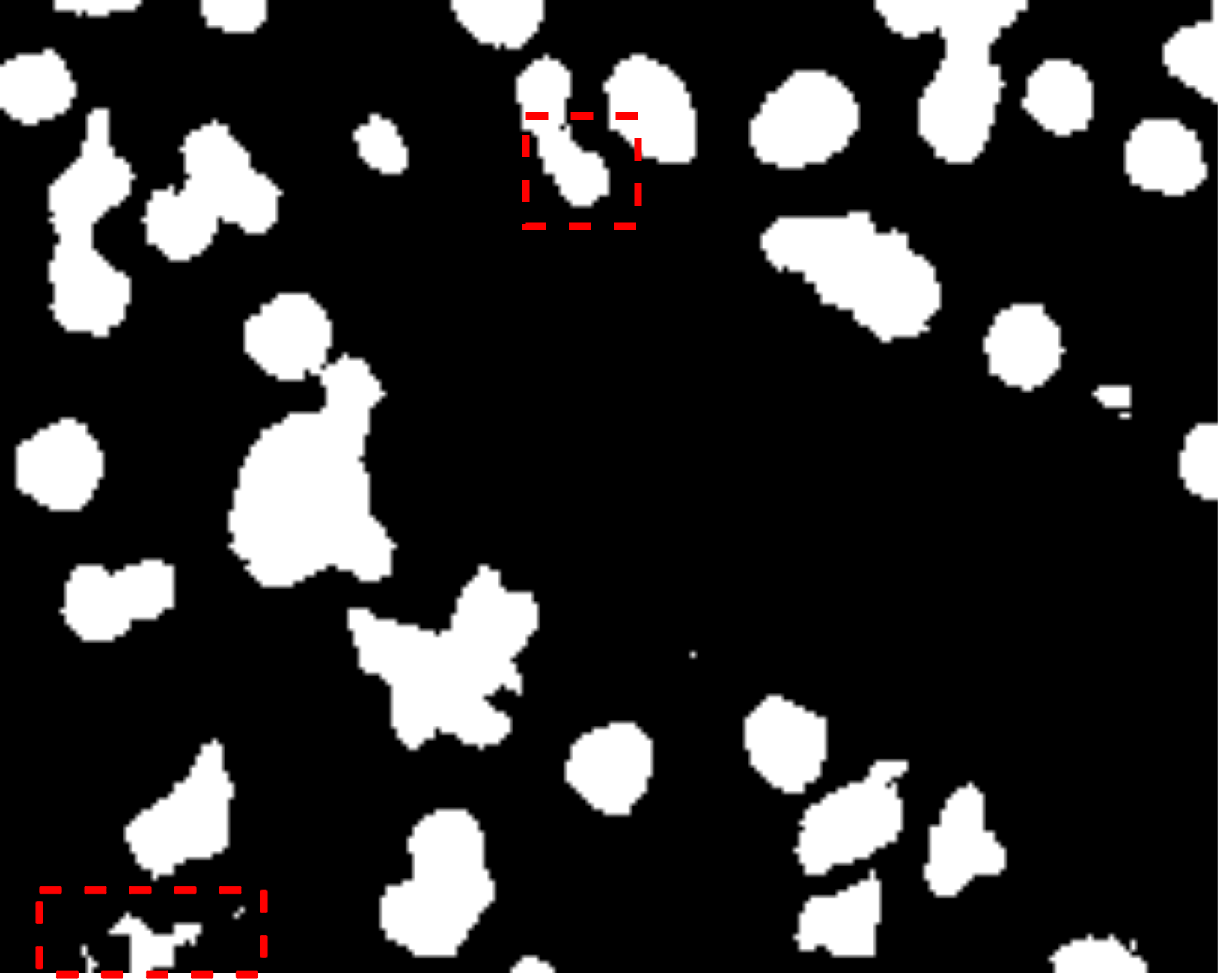} &
\includegraphics[height=0.23\linewidth, width=0.23\linewidth]{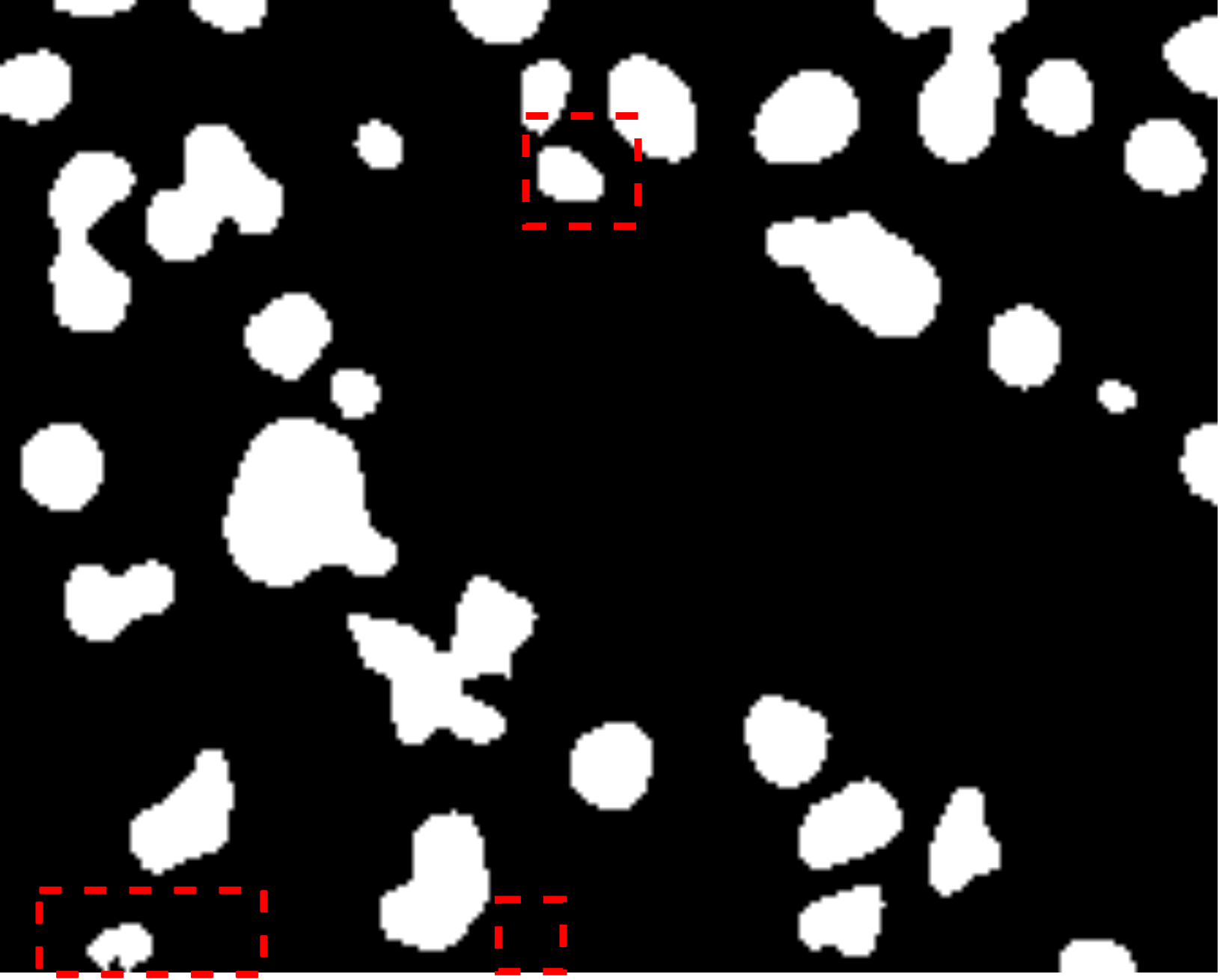} &
\includegraphics[height=0.23\linewidth, width=0.23\linewidth]{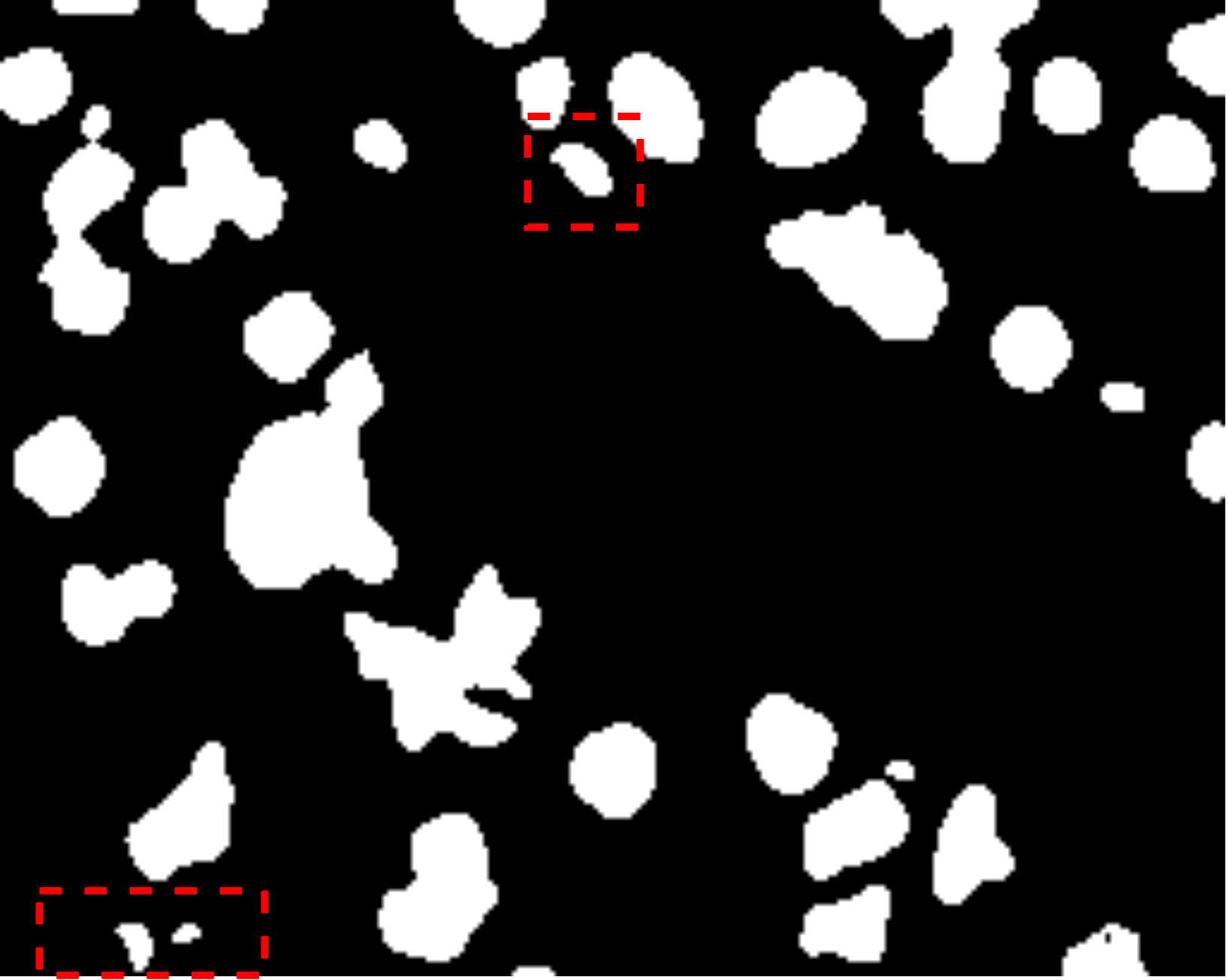} &
\includegraphics[height=0.23\linewidth, width=0.23\linewidth]{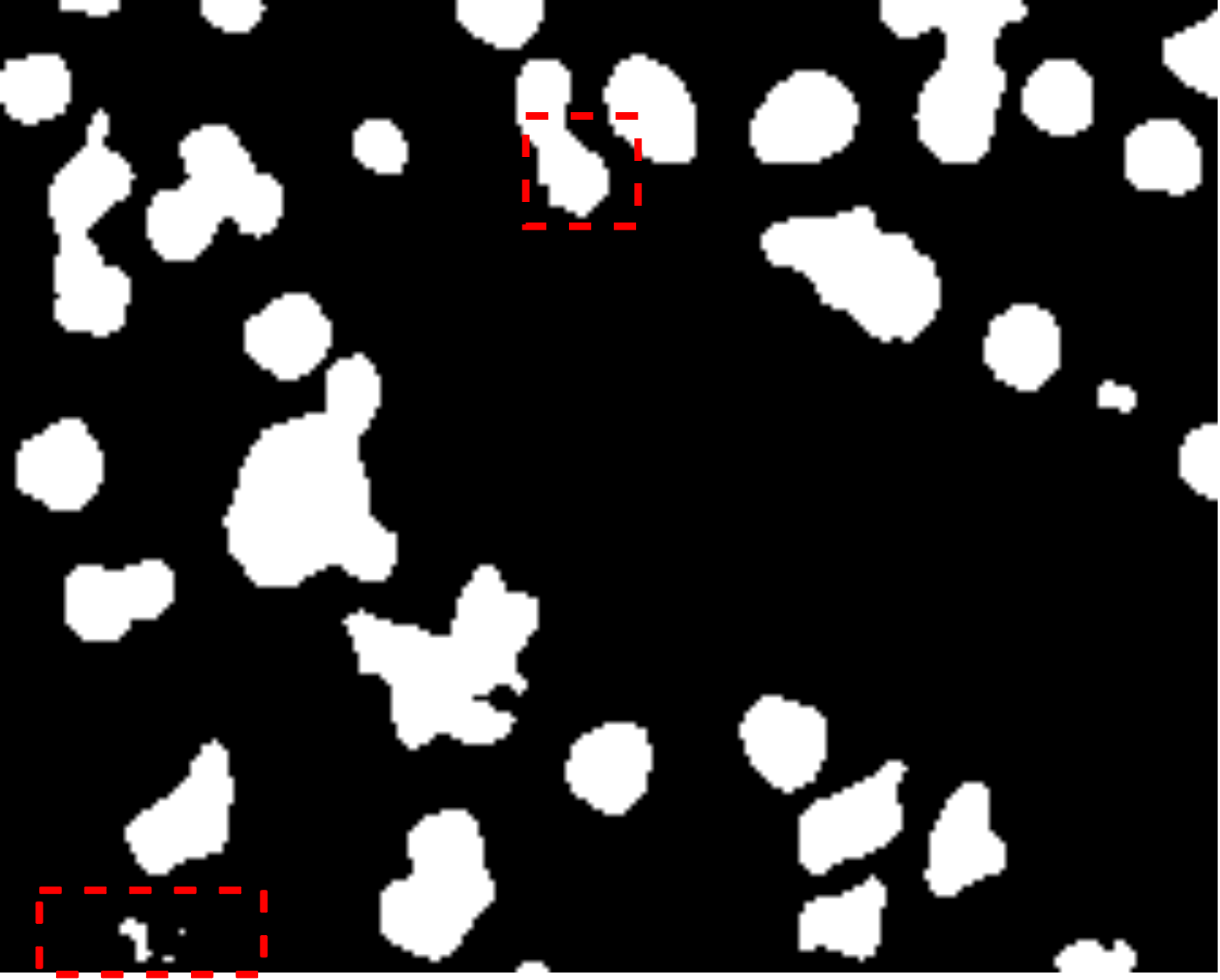} \\

(e) Swin-Unet & (f) MET-Net & (g) TransUNet & (h) FSCA-Net \\

\end{tabular}
\caption{The comparison of visualization prediction between QTSeg and other methods on the DSB2018 dataset. The red dash bounding boxes
represent the improved segmentation by QTSeg, and the yellow dash bounding boxes show their existing shortcomings.}
\label{fig:dsb2018_viz}
\end{figure}

Fig.~\ref{fig:dsb2018_viz} visually illustrates the segmentation performance comparison between our proposed method and other high-scoring approaches. Notably, our method successfully classifies a background area that is similar to the cell object, as evidenced by the first red bounding box, a task that other methods struggle with. This accurate classification in the first red bounding box highlights the effectiveness of our DMA in integrating global information into the decoder, providing crucial contextual information for mask decoding. Moreover, even in architectures such as TranUNet and Swin-Unet, known for handling long-range dependencies, the first red bounding box is identified as a cell, further showcasing the efficiency of our DMA module above these approaches.

On the other hand, while CNN-based models excel in capturing information in the second red bounding box, transformer-based models seem to face challenges in this area, as depicted in Fig.~\ref{fig:dsb2018_viz}e-h. Nonetheless, our model exhibits these limitations, as indicated by the yellow box, where improvements are needed compared to the labeled image. This constraint arises from the presence of blur and small objects in the image, requiring further enhancement in the future.

\subsubsection{Results on Breast Cancer Segmentation}

\begin{figure}
\centering
\fontsize{9pt}{9pt}\selectfont
\begin{tabular}{cccc}

\includegraphics[height=0.23\linewidth, width=0.23\linewidth]{figures/results/BUSI/FigBUSI_m108.png} &
\includegraphics[height=0.23\linewidth, width=0.23\linewidth]{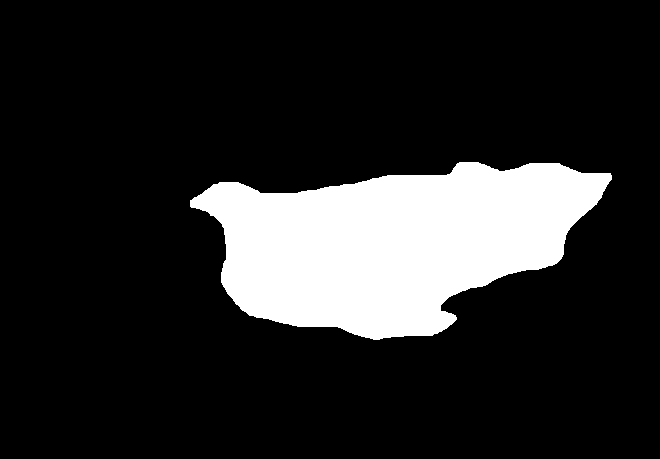} &
\includegraphics[height=0.23\linewidth, width=0.23\linewidth]{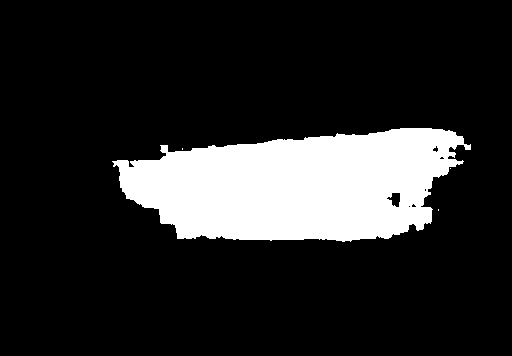} &
\includegraphics[height=0.23\linewidth, width=0.23\linewidth]{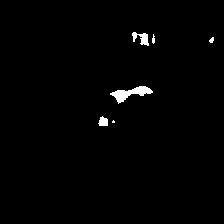} \\

(a) Inputs & (b) Ground Truth & (c) Ours & (d) FSCA-Net \\
\includegraphics[height=0.23\linewidth, width=0.23\linewidth]{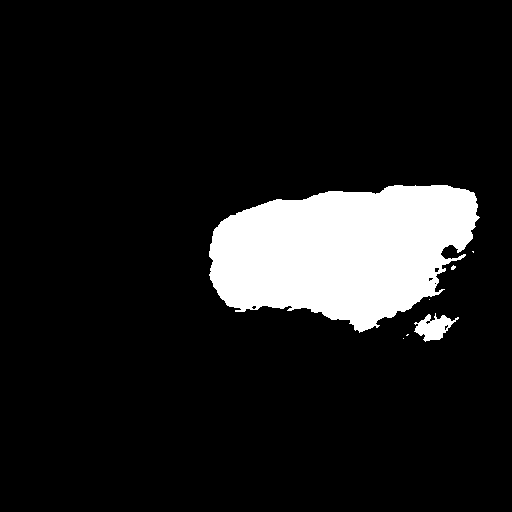} &
\includegraphics[height=0.23\linewidth, width=0.23\linewidth]{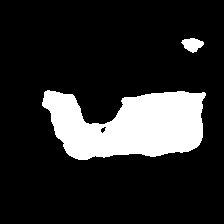} &
\includegraphics[height=0.23\linewidth, width=0.23\linewidth]{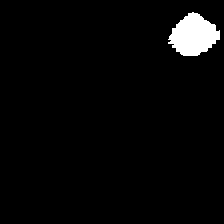} &
\includegraphics[height=0.23\linewidth, width=0.23\linewidth]{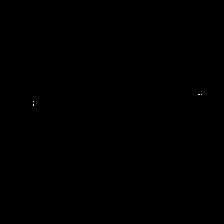} \\

(d) MHorUNet & (e) TransUNet & (f) MISSFormer & (g) U-Net \\

\end{tabular}
\caption{Visualization of a failure case 
% for the Malignant sample (108) 
from the BUSI dataset.}
\label{fig:busi_viz}
\end{figure}

As depicted in Table~\ref{tab:lesion_performance}, QTSeg demonstrates superior performance across all metrics when evaluated using a five-fold assessment. In particular, our QTSeg model achieves remarkable results with 82.09\% Dice and 73.85\% IoU, outperforming FSCA-Net by 0.84\% and 0.90\%, respectively. In Fig.~\ref{fig:busi_viz}, a failure case study on the BUSI dataset showcases QTSeg's accurate predictions compared to other methods. Despite challenging features in the sample, QTSeg excels in segmenting a portion of the sample accurately, unlike models like U-Net~\cite{unet} and MISSFormer~\cite{missformer} that struggle with the segmentation task. While approaches like MHorUNet~\cite{mhorunet} and TransUNet~\cite{transunet} can predict the target area, our QTSeg model outperforms them by generating predictions that closely match the ground truth with significantly less error. This visualization effectively illustrates how the DMA module leverages global information to enhance its predictions, especially when dealing with similar features extracted by the encoder model. In Fig.~\ref{fig:busi_viz}a, the input image lacks distinct features for differentiation, leading to segmentation areas that closely resemble the background, as evident in the misclassification shown in Fig.~\ref{fig:busi_viz}g of a CNN-based method. In contrast, employing a similar approach to the CNN-based method, QTSeg successfully discriminates between the background and foreground regions by incorporating information from other areas. This exemplifies how the DMA module effectively addresses the challenges related to local-global information integration in image segmentation tasks.

\setlength{\tabcolsep}{2pt}

\subsubsection{Results on Retinal Vessel Segmentation}

It is crucial to assess our method's performance on high-resolution images and various segmentation tasks. Table~\ref{tab:fives_performance} provides a comparison of our approach with other recent methods using Acc, Dice, and IoU metrics. Once more, QTSeg stands out by achieving the highest scores across all metrics: 98.93\% for Acc, 90.37\% for Dice, and 84.26\% for IoU. It is evident that QTSeg excels in accurately segmenting complex structures and nuances present in high-resolution images. The consistently high scores in Acc, Dice, and IoU metrics signify the robustness and reliability of our approach in capturing fine details and boundaries within the images. Furthermore, the superior performance on vessel segmentation tasks highlights the versatility and effectiveness of QTSeg across different segmentation scenarios. Overall, these results underscore the potential of our method as a leading solution in image segmentation tasks.

\begin{table}
\centering
\caption{Performance comparison between QTSeg and other state-of-the-art methods on the Retinal Vessel Segmentation dataset.}
\label{tab:fives_performance}
\setlength{\tabcolsep}{3pt}
\begin{tabular}{l|ccc}
\hline
 \textbf{Method}                                        & \textbf{Acc}$\uparrow$     & \textbf{Dice}$\uparrow$    & \textbf{IoU}$\uparrow$\\
\hline
U-Net (2015)~\cite{unet}                                & 98.66             & 88.87             & 80.77             \\
% R2U-Net ()~\cite{?}                                   & 98.09             & 84.92             & 74.65             \\
% GCN ()~\cite{?}                                       & 98.79             & 90.02             & 82.60             \\
% ENet ()~\cite{?}                                      & 98.67             & 89.09             & 81.10             \\
PSPNet (2017)~\cite{zhao_hengshuang2017pspnet}          & 98.78             & 89.88             & 82.35             \\
SegNet (2017)~\cite{badrinarayanan_vijay2017segnet}     & 98.13             & 85.09             & 74.98             \\
Attention-Unet (2018)~\cite{oktay2018attentionunet}     & 98.68             & 88.81             & 80.73             \\
Deeplab V3+ (2018)~\cite{liangchieh2018deeplabv3plus}   & 98.50             & 88.56             & 80.75             \\
CBAM (2018)~\cite{woo2018cbam}                          & 98.67             & 88.50             & 80.29             \\
SK (2019)~\cite{xiang2019sk_cvpr}                       & 98.58             & 88.35             & 79.94             \\
SGAT-Net (2023)~\cite{ji_lin2023sgaf}                   & \underline{98.86}    & \underline{90.51}    & \underline{83.47}    \\
Swin-Unet (2023)~\cite{swinunet}                        & 98.82             & 90.13             & 82.76             \\
TransUNet (2024)~\cite{transunet}                       & 98.83             & 90.37             & 83.17             \\
\hline
\textbf{QTSeg (Ours) }                                  & \textbf{98.93}    & \textbf{90.98}    & \textbf{84.26}             \\

\hline
\end{tabular}
\end{table}

\begin{figure*}[!t]
\setlength{\tabcolsep}{2pt}
\centering
\fontsize{9pt}{9pt}\selectfont
\begin{tabular}{ccccc}

\includegraphics[height=0.2\linewidth, width=0.2\linewidth]{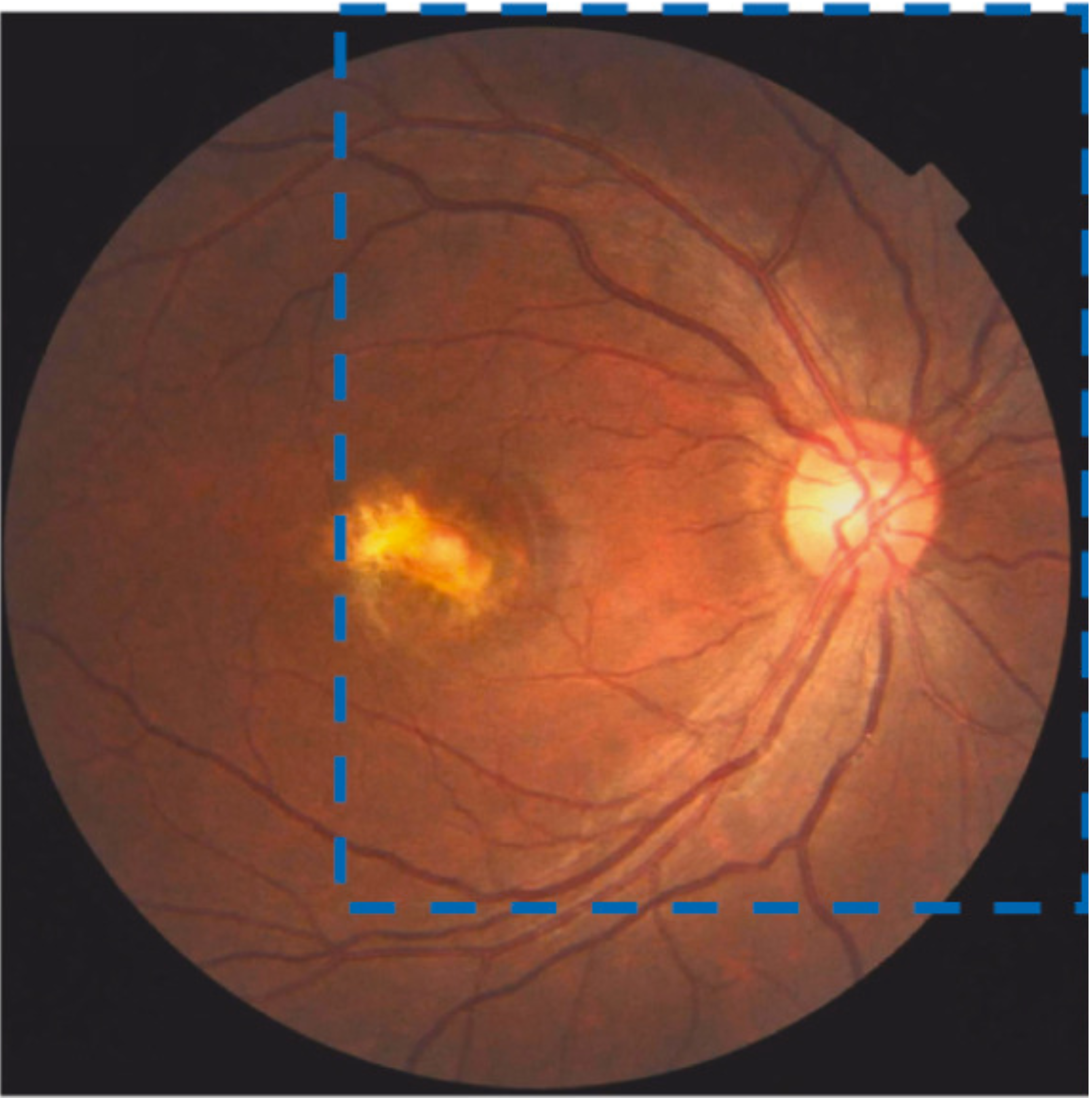} &
\includegraphics[height=0.2\linewidth, width=0.2\linewidth]{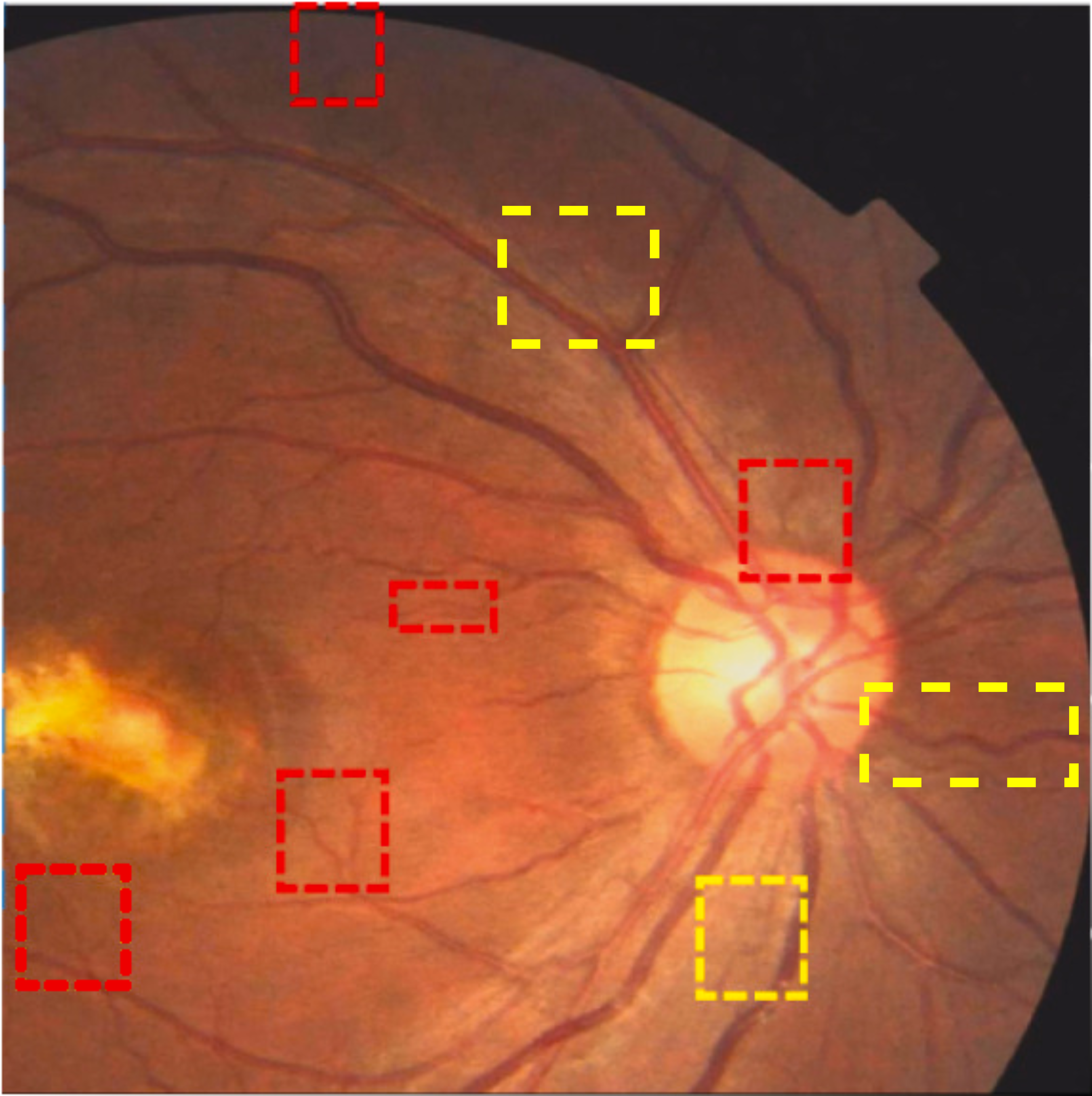} &
\includegraphics[height=0.2\linewidth, width=0.2\linewidth]{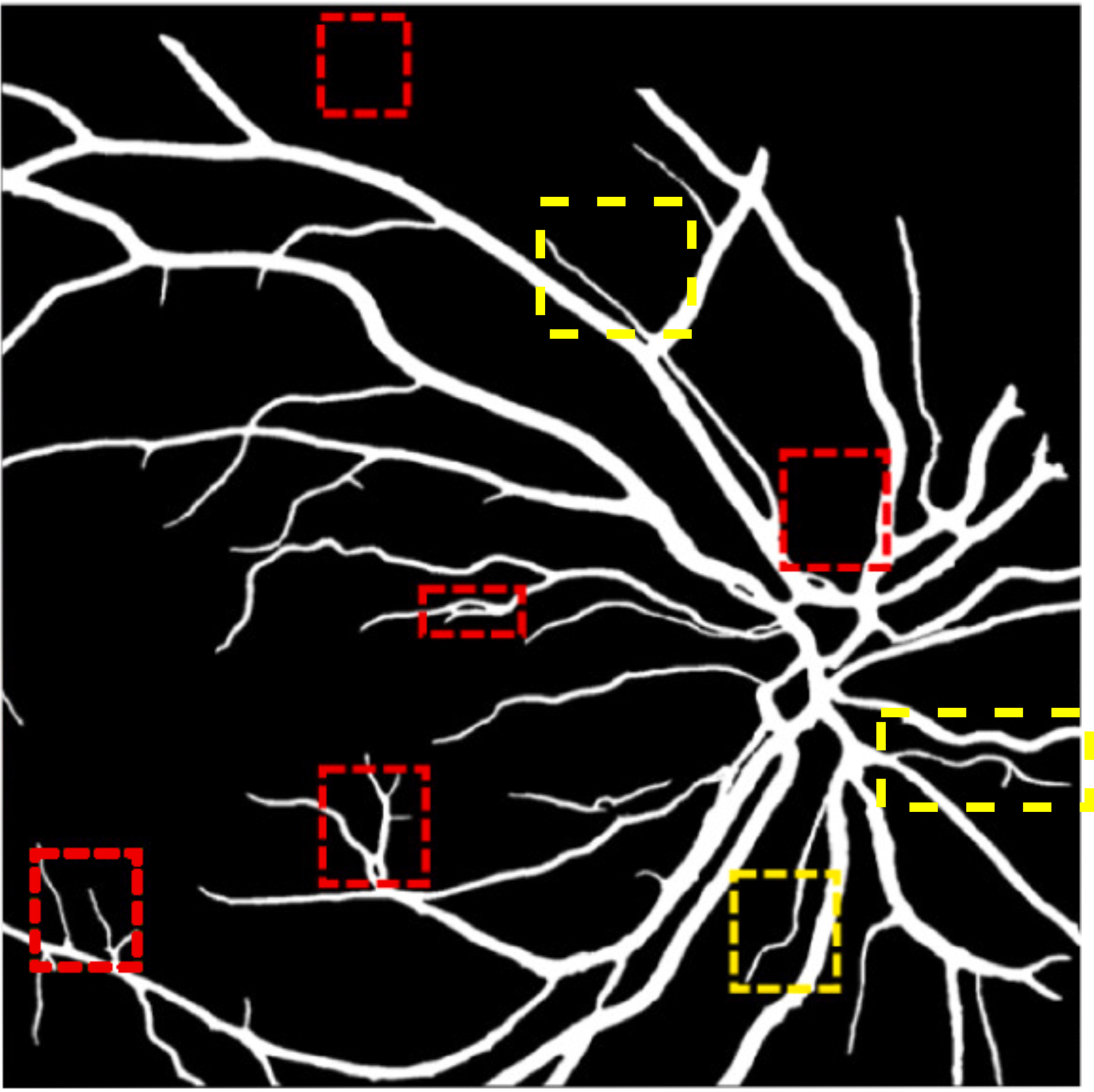} &
\includegraphics[height=0.2\linewidth, width=0.2\linewidth]{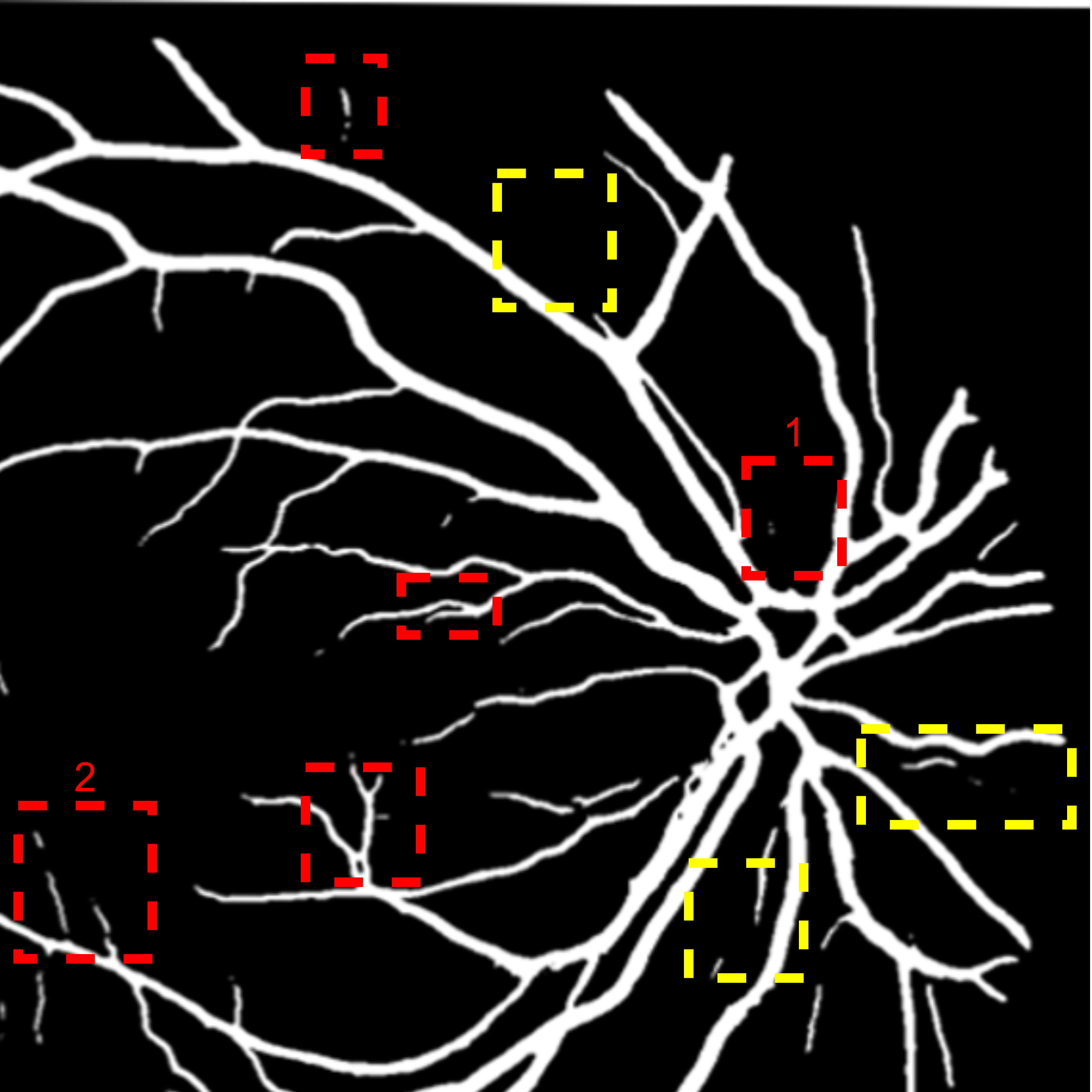} &
\includegraphics[height=0.2\linewidth, width=0.2\linewidth]{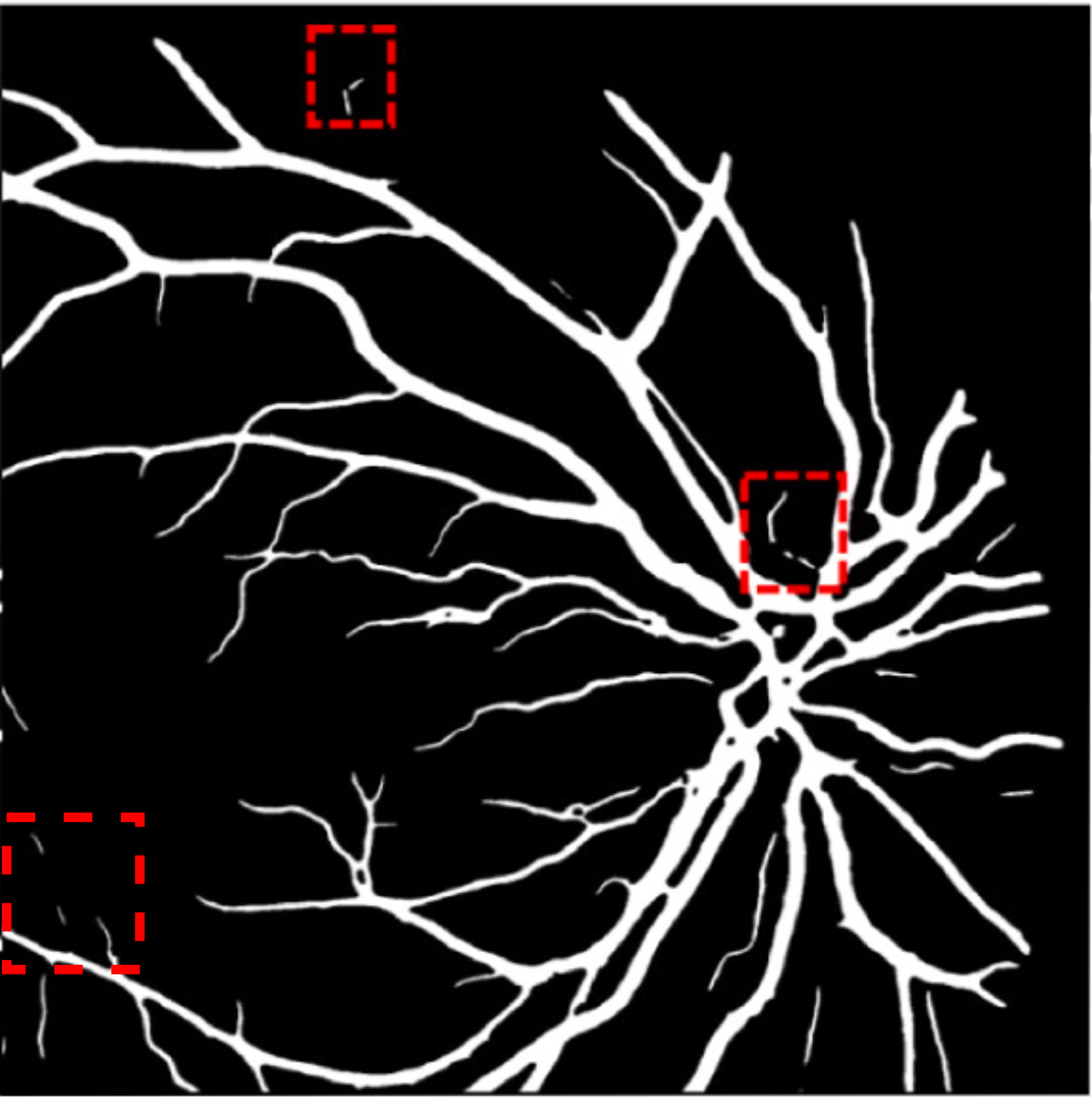} \\
(a) Input & (b) Input crop & (c) Ground Truth & (d) Ours & (e) U-Net \\

\includegraphics[height=0.2\linewidth, width=0.2\linewidth]{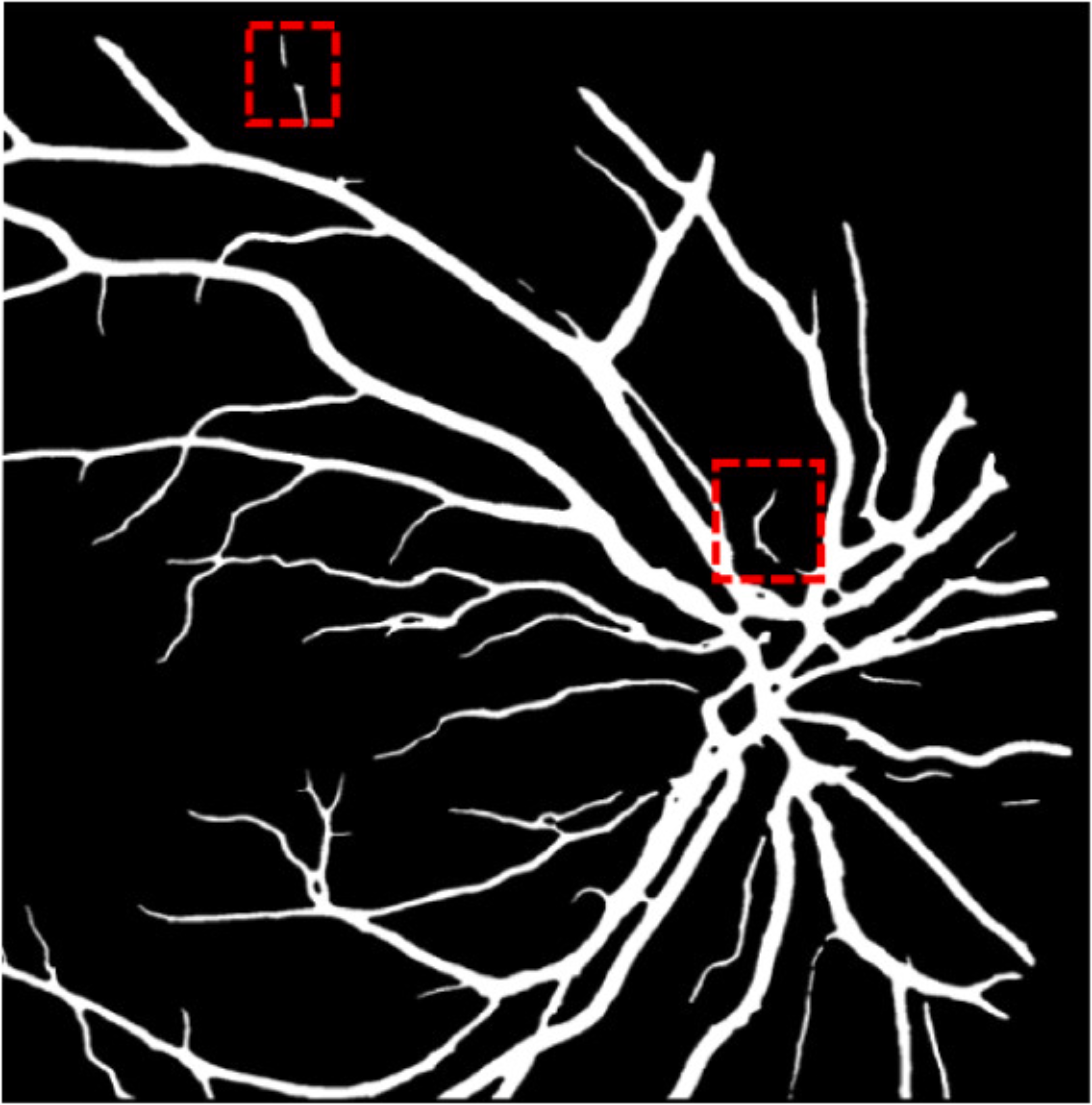} &
\includegraphics[height=0.2\linewidth, width=0.2\linewidth]{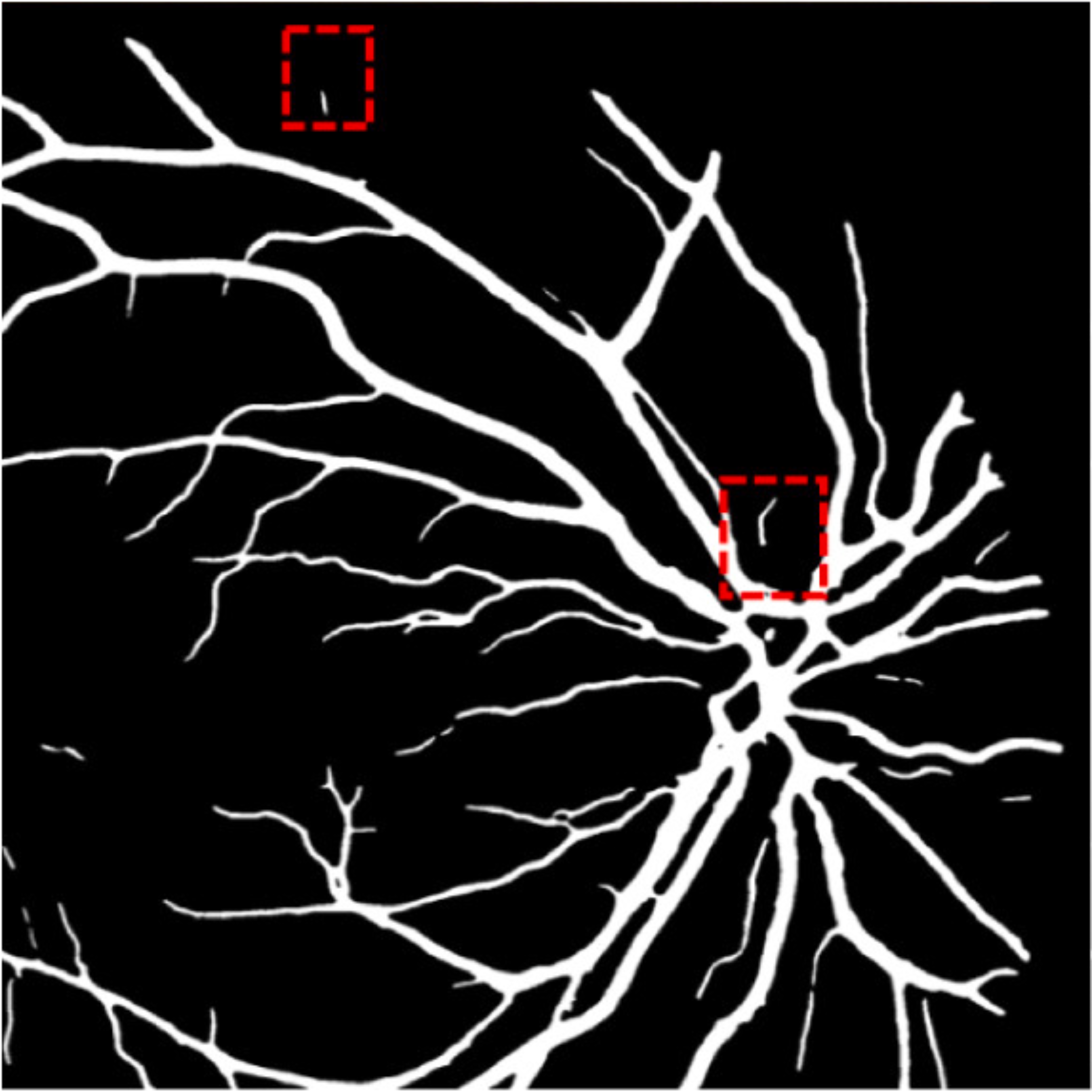} &
\includegraphics[height=0.2\linewidth, width=0.2\linewidth]{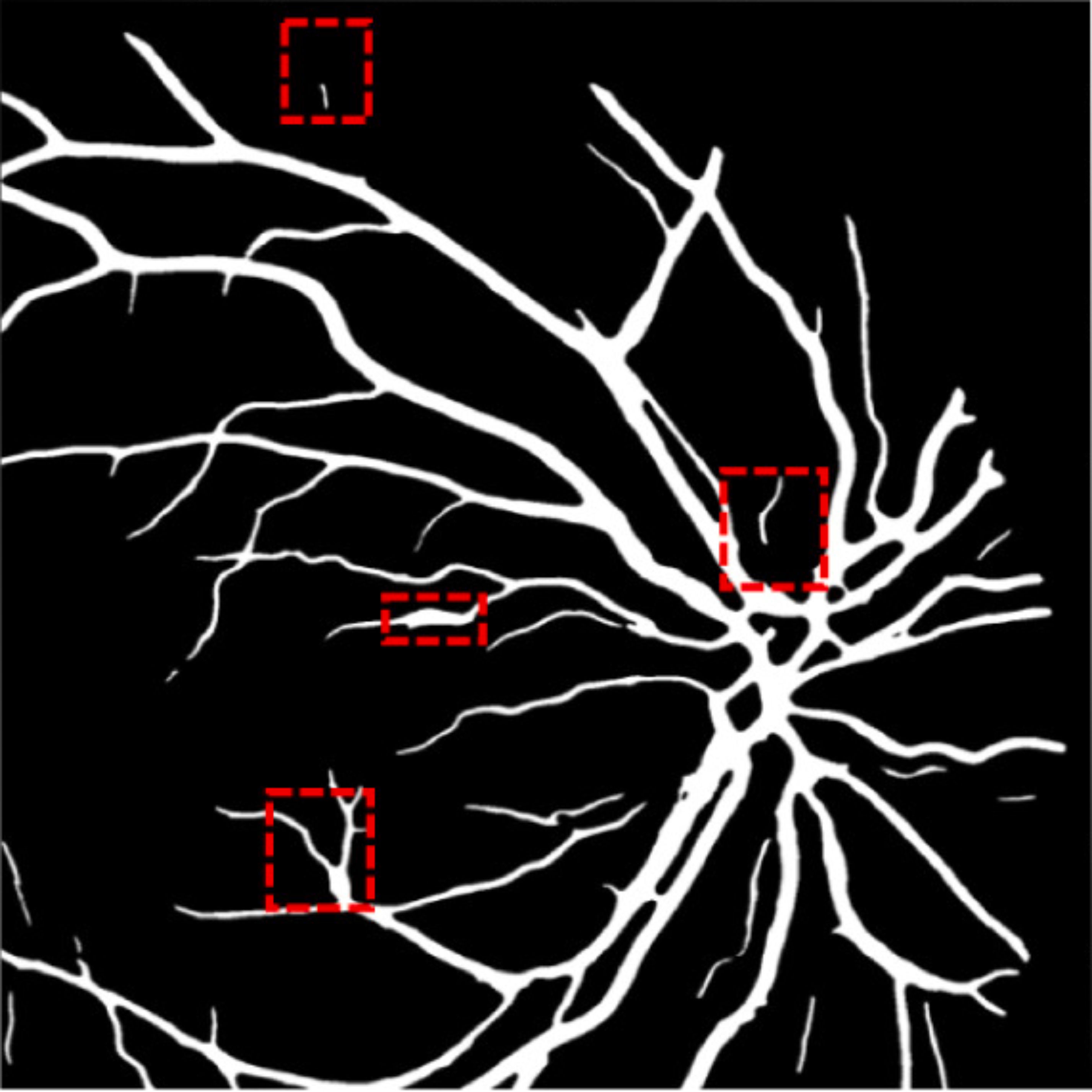} &
\includegraphics[height=0.2\linewidth, width=0.2\linewidth]{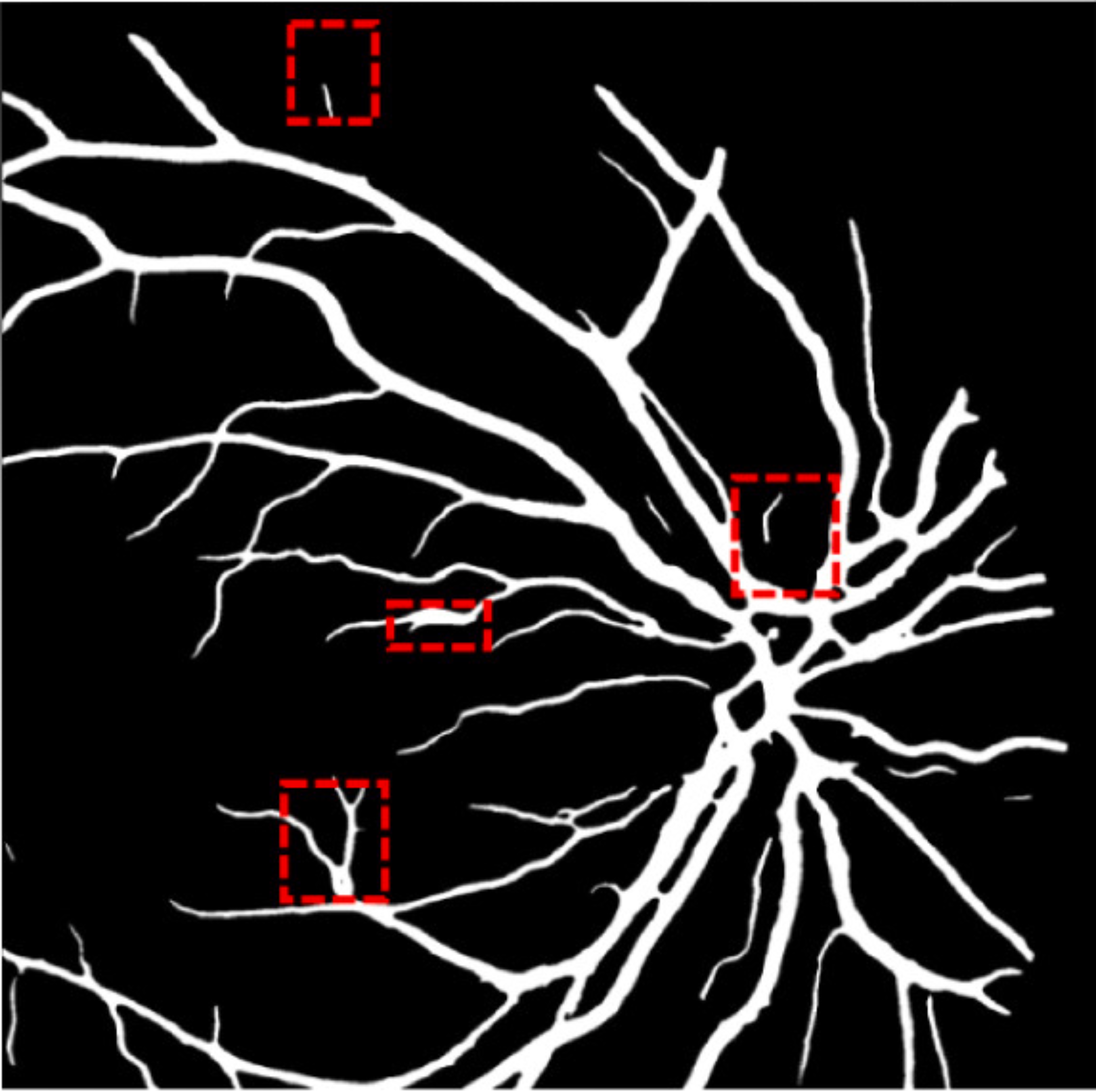} &
\includegraphics[height=0.2\linewidth, width=0.2\linewidth]{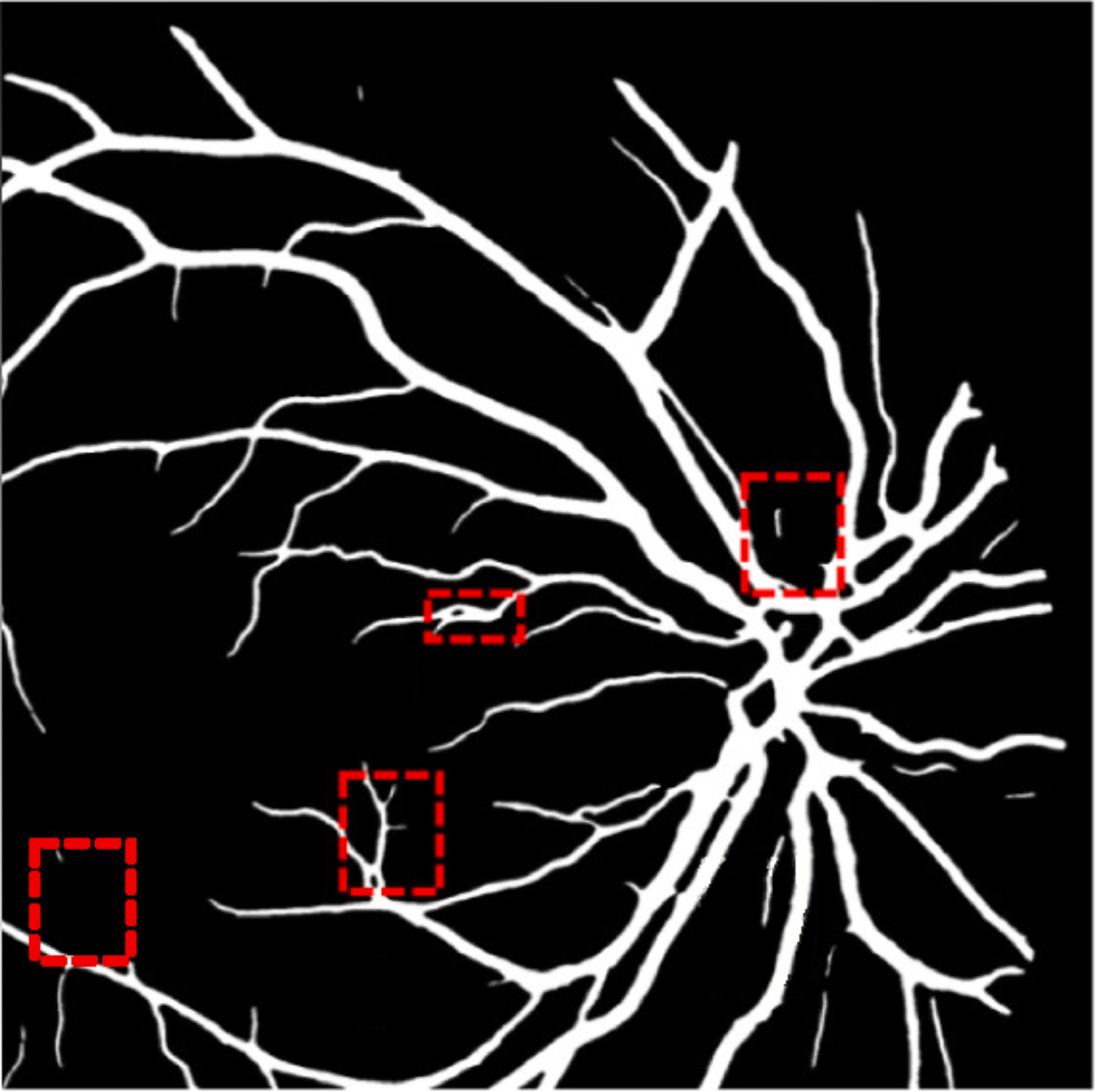} \\ 

(f) Attention-Unet & (g) Deeplab V3+ & (h) SK & (i) TransUNet & (j) SGAT-Net \\

\end{tabular}
\centering
\caption{The comparison of visualization prediction between QTSeg and other methods on the FIVES dataset. The red dash bounding boxes represent the improved segmentation by QTSeg, and the yellow dash bounding boxes show their existing shortcomings.}
\label{fig:fives_viz}
\end{figure*}

A deep visualization analysis is presented in Fig.~\ref{fig:fives_viz} for a deeper understanding of our model performance. The red and yellow dash bounding boxes show the improvement and the shortcomings of the proposed method, respectively. The QTSeg can distinguish the blood vessel from the background and does not misclassify the vessel according to the second red bounding box in the figure. Besides, the first red bounding box, which has features similar to those of the blood vessel, has a good classification compared to other methods. However, our method still gains inferior performance on some blurred blood vessels or similar vessels, as highlighted in the yellow bounding boxes. The design of the DMA module successfully addresses the challenges posed by local and global information constraints in both CNN and Transformer architectures. The visual representation in the figure illustrates how our method minimizes mask errors while also being computationally efficient and parameter-light compared to alternative approaches. Moreover, the masks produced by QTSeg demonstrate a remarkable ability to differentiate between similar features like colors and shapes, enabling accurate inferences based on the blood vessel structures. This capability is exemplified by the distinct delineation of blood vessels and lesions within the first red bounding box in the figure. These observations emphasize the significance of integrating transformer mechanisms with CNN feature extraction, enabling our model to leverage global information effectively.

\begin{table*}
\centering
\caption{Ablation studies on ISIC2016 dataset.}
\label{tab:ablation}
\setlength{\tabcolsep}{3pt}
\begin{tabular}{ccccc|ccccrr}
\hline  
&\multicolumn{4}{c|}{\textbf{Components}}   &\multirow{2}{*}{\textbf{MAE}$\downarrow$}&\multirow{2}{*}{\textbf{Acc}$\uparrow$}&\multirow{2}{*}{\textbf{Dice}$\uparrow$}&\multirow{2}{*}{\textbf{IoU}$\uparrow$} &\multirow{2}{*}{\textbf{Params}$\downarrow$}&\multirow{2}{*}{\textbf{FLOPs}$\downarrow$}   \\
\textbf{Setting} & \textbf{Method}    & \textbf{Encoder}    &  \textbf{Decoder}             &   \textbf{MLFD}    &          &            &            &          &          \\
\hline
Baseline&  MedSAM   & ViT Base   & MedSAM decoder       &   -       & -          & -        & -          & -          & 93.74 M & 488.24 G    \\
1       &  MedSAM   & TinyViT    & MedSAM decoder       &   -       & 3.88       & 96.12    & 92.33      & 86.59      & 9.79 M  & 39.91 G     \\
2       &  MedSAM   & CNN-based  & MedSAM decoder       &   -       & 4.00       & 96.00    & 92.14      & 86.31      & \textbf{4.63 M}  & \textbf{1.89 G }     \\
3       &  QTSeg    & CNN-based  & Multi-MedSAM decoder &   -       & 3.87       & 96.13    & 91.92      & 85.87      & 9.41 M  & \underline{2.19} G      \\
4       &  QTSeg    & CNN-based  & Multi-MedSAM decoder & \checkmark& 3.59       & 96.41    & \underline{92.42}      & \underline{86.74}      & 9.69 M  & 2.29 G      \\
5       &  QTSeg    & CNN-based  & Multi-DMAD decoder   &   -       & \underline{3.07}       & \underline{96.93}    & 92.06      & 86.30      & \underline{8.94 M}  & 2.49 G      \\
\hline
6       &  QTSeg    & CNN-based  & Multi-DMAD decoder   & \checkmark& \textbf{3.00}       & \textbf{97.01 }   & \textbf{92.45}      & \textbf{86.92}      & 9.21 M  & 2.60 G      \\
\hline
\end{tabular}
\end{table*}

\subsection{Ablation Studies}
\label{ablation}

To evaluate the effectiveness of our architecture, we conducted various ablation studies on the QTSeg architecture. Due to limitations in resources, we chose not to assess the baseline MedSAM~\cite{medsam} with the ViT base architecture. Instead, we replaced the ViT base with TinyViT~\cite{wu_kan2022tinyvit} to evaluate the model's performance within the MedSAM framework, as shown in Setting~1 of Table~\ref{tab:ablation}. This configuration aligns MedSAM with the architecture of MobileSAM~\cite{mobilesam}, known for its strong performance in general segmentation tasks. Additionally, we replaced TinyViT with a CNN-based encoder model without modifying the MedSAM architecture (Setting~2). A comparison between Settings 1 and 2 reveals that transitioning from TinyViT to the CNN-based encoder results in a decrease in both model accuracy and complexity. To enhance model performance to meet a similar baseline standard, we utilize the U-Shape architecture detailed in Section~\ref{sec:method} and showcased it in Settings~3,~4, ~5, and~6. Settings~3 and~4 illustrate the use of multiple decoders from the MedSAM decoder with and without our MLFD modules, emphasizing the significance of MLFD in enhancing model performance by distributing features across multiple stages. Finally, by replacing the MedSAM decoder with our proposed DMAD decoder, we achieved superior performance in Settings~5 and~6. Notably, with parameters similar to Setting~1, our QTSeg demonstrated a significant reduction in model complexity from 39.91 GFLOPs to 2.60 GFLOPs while maintaining or even surpassing performance metrics such as MAE, Acc, and IoU.

\section{Conclusion}
\label{sec:conclusion}
In this study, we introduce QTSeg, a query token-based approach for medical image segmentation incorporating a mask decoder with DMAD. The proposed DMAD captures both local and global contexts within the level features of the CNN-based encoder. Additionally, we introduce an adaptive MLFD module to efficiently distribute features across multiple stages, thereby enhancing the overall performance of our models. Our research showcases that our proposed architecture strikes a balance between model complexity and segmentation performance in various medical image segmentation tasks like poly, lesion, breast cancer, cell, and retina vessel segmentation. However, our method encounters challenges with small and blue objects, indicating areas for improvement in the future. 

\bibliographystyle{IEEEtran}
\bibliography{BibTeX}

% Generated by IEEEtran.bst, version: 1.14 (2015/08/26)
\begin{thebibliography}{10}
\providecommand{\url}[1]{#1}
\csname url@samestyle\endcsname
\providecommand{\newblock}{\relax}
\providecommand{\bibinfo}[2]{#2}
\providecommand{\BIBentrySTDinterwordspacing}{\spaceskip=0pt\relax}
\providecommand{\BIBentryALTinterwordstretchfactor}{4}
\providecommand{\BIBentryALTinterwordspacing}{\spaceskip=\fontdimen2\font plus
\BIBentryALTinterwordstretchfactor\fontdimen3\font minus \fontdimen4\font\relax}
\providecommand{\BIBforeignlanguage}[2]{{%
\expandafter\ifx\csname l@#1\endcsname\relax
\typeout{** WARNING: IEEEtran.bst: No hyphenation pattern has been}%
\typeout{** loaded for the language `#1'. Using the pattern for}%
\typeout{** the default language instead.}%
\else
\language=\csname l@#1\endcsname
\fi
#2}}
\providecommand{\BIBdecl}{\relax}
\BIBdecl

\bibitem{xuyan2024advancesMIS}
Y.~Xu, R.~Quan, W.~Xu, Y.~Huang, X.~Chen, and F.~Liu, ``Advances in medical image segmentation: A comprehensive review of traditional, deep learning and hybrid approaches,'' \emph{Bioengineering}, vol.~11, no.~10, 2024.

\bibitem{rahman2023MISviaCAD}
M.~M. Rahman and R.~Marculescu, ``Medical image segmentation via cascaded attention decoding,'' in \emph{2023 IEEE/CVF Winter Conference on Applications of Computer Vision (WACV)}, 2023, pp. 6211--6220.

\bibitem{morra2021icpr}
L.~Morra, L.~Piano, F.~Lamberti, and T.~Tommasi, ``Bridging the gap between natural and medical images through deep colorization,'' in \emph{2020 25th International Conference on Pattern Recognition (ICPR)}, 2021, pp. 835--842.

\bibitem{kimhee2022transfer}
H.~E. Kim, A.~Cosa-Linan, N.~Santhanam, M.~Jannesari, M.~E. Maros, and T.~Ganslandt, ``Transfer learning for medical image classification: a literature review,'' \emph{BMC medical imaging}, vol.~22, no.~1, p.~69, 2022.

\bibitem{guan2022ieeetbe}
H.~Guan and M.~Liu, ``Domain adaptation for medical image analysis: A survey,'' \emph{IEEE Transactions on Biomedical Engineering}, vol.~69, no.~3, pp. 1173--1185, 2022.

\bibitem{h2former}
A.~He, K.~Wang, T.~Li, C.~Du, S.~Xia, and H.~Fu, ``{H2Former: An Efficient Hierarchical Hybrid Transformer for Medical Image Segmentation},'' \emph{IEEE Transactions on Medical Imaging}, vol.~42, no.~9, pp. 2763--2775, 2023.

\bibitem{missformer}
X.~Huang, Z.~Deng, D.~Li, X.~Yuan, and Y.~Fu, ``{MISSFormer: An Effective Transformer for 2D Medical Image Segmentation},'' \emph{IEEE Transactions on Medical Imaging}, vol.~42, no.~5, pp. 1484--1494, 2023.

\bibitem{unet}
O.~Ronneberger, P.~Fischer, and T.~Brox, ``{U-Net: Convolutional Networks for Biomedical Image Segmentation},'' in \emph{Medical Image Computing and Computer-Assisted Intervention -- MICCAI 2015}, N.~Navab, J.~Hornegger, W.~M. Wells, and A.~F. Frangi, Eds.\hskip 1em plus 0.5em minus 0.4em\relax Cham: Springer International Publishing, 2015, pp. 234--241.

\bibitem{unet++}
Z.~Zhou, M.~M.~R. Siddiquee, N.~Tajbakhsh, and J.~Liang, ``{UNet++: Redesigning Skip Connections to Exploit Multiscale Features in Image Segmentation},'' \emph{IEEE Transactions on Medical Imaging}, vol.~39, no.~6, pp. 1856--1867, 2020.

\bibitem{isensee2021nnunet}
F.~Isensee, P.~F. Jaeger, S.~A. Kohl, J.~Petersen, and K.~H. Maier-Hein, ``{nnU-Net: a self-configuring method for deep learning-based biomedical image segmentation},'' \emph{Nature methods}, vol.~18, no.~2, pp. 203--211, 2021.

\bibitem{rahman2024mist}
M.~M. Rahman, S.~Shokouhmand, S.~Bhatt, and M.~Faezipour, ``Mist: Medical image segmentation transformer with convolutional attention mixing (cam) decoder,'' in \emph{2024 IEEE/CVF Winter Conference on Applications of Computer Vision (WACV)}, 2024, pp. 403--412.

\bibitem{vaswani2017attention}
A.~Vaswani \emph{et~al.}, ``{Attention is all you need},'' in \emph{Proceedings of the 31st International Conference on Neural Information Processing Systems}, ser. NIPS'17.\hskip 1em plus 0.5em minus 0.4em\relax Red Hook, NY, USA: Curran Associates Inc., 2017, p. 6000–6010.

\bibitem{kelei2023transformerMIA}
K.~He, C.~Gan, Z.~Li, I.~Rekik, Z.~Yin, W.~Ji, Y.~Gao, Q.~Wang, J.~Zhang, and D.~Shen, ``Transformers in medical image analysis,'' \emph{Intelligent Medicine}, vol.~3, no.~1, pp. 59--78, 2023.

\bibitem{transunet}
J.~Chen \emph{et~al.}, ``{TransUNet: Rethinking the U-Net architecture design for medical image segmentation through the lens of transformers},'' \emph{Medical Image Analysis}, p. 103280, 2024.

\bibitem{swinunet}
H.~Cao \emph{et~al.}, ``{Swin-Unet: Unet-Like Pure Transformer for Medical Image Segmentation},'' in \emph{Computer Vision -- ECCV 2022 Workshops}, L.~Karlinsky, T.~Michaeli, and K.~Nishino, Eds.\hskip 1em plus 0.5em minus 0.4em\relax Cham: Springer Nature Switzerland, 2023, pp. 205--218.

\bibitem{honglin2024cmlformer}
H.~Wu, M.~Zhang, P.~Huang, and W.~Tang, ``Cmlformer: Cnn and multiscale local-context transformer network for remote sensing images semantic segmentation,'' \emph{IEEE Journal of Selected Topics in Applied Earth Observations and Remote Sensing}, vol.~17, pp. 7233--7241, 2024.

\bibitem{xian2023lighterbetter}
X.~Lin, L.~Yu, K.-T. Cheng, and Z.~Yan, ``The lighter the better: Rethinking transformers in medical image segmentation through adaptive pruning,'' \emph{IEEE Transactions on Medical Imaging}, vol.~42, no.~8, pp. 2325--2337, 2023.

\bibitem{sanghyun2018cbam}
S.~Woo, J.~Park, J.-Y. Lee, and I.~S. Kweon, ``Cbam: Convolutional block attention module,'' in \emph{Proceedings of the European conference on computer vision (ECCV)}, 2018, pp. 3--19.

\bibitem{chunfu2021cross_attention}
C.-F.~R. Chen, Q.~Fan, and R.~Panda, ``{Crossvit: Cross-attention multi-scale vision transformer for image classification},'' in \emph{Proceedings of the IEEE/CVF international conference on computer vision}, 2021, pp. 357--366.

\bibitem{dense_unet}
X.~Li, H.~Chen, X.~Qi, Q.~Dou, C.-W. Fu, and P.-A. Heng, ``{H-DenseUNet: Hybrid Densely Connected UNet for Liver and Tumor Segmentation From CT Volumes},'' \emph{IEEE Transactions on Medical Imaging}, vol.~37, no.~12, pp. 2663--2674, 2018.

\bibitem{attention_unet}
O.~Oktay \emph{et~al.}, ``{Attention u-net: Learning where to look for the pancreas},'' \emph{arXiv preprint arXiv:1804.03999}, 2018.

\bibitem{kamnitsas2017deepmedic}
K.~Kamnitsas, C.~Ledig, V.~F. Newcombe, J.~P. Simpson, A.~D. Kane, D.~K. Menon, D.~Rueckert, and B.~Glocker, ``Efficient multi-scale 3d cnn with fully connected crf for accurate brain lesion segmentation,'' \emph{Medical Image Analysis}, vol.~36, pp. 61--78, 2017.

\bibitem{myronenko2019segresnet}
A.~Myronenko, ``3d mri brain tumor segmentation using autoencoder regularization,'' in \emph{Brainlesion: Glioma, Multiple Sclerosis, Stroke and Traumatic Brain Injuries: 4th International Workshop, BrainLes 2018, Held in Conjunction with MICCAI 2018, Granada, Spain, September 16, 2018, Revised Selected Papers, Part II 4}.\hskip 1em plus 0.5em minus 0.4em\relax Springer, 2019, pp. 311--320.

\bibitem{cv_task1}
Q.~Zhang and Y.-B. Yang, ``{Rest: An efficient transformer for visual recognition},'' \emph{Advances in neural information processing systems}, vol.~34, pp. 15\,475--15\,485, 2021.

\bibitem{swintexco}
D.~T. Tran \emph{et~al.}, ``{SwinTExCo: Exemplar-based video colorization using Swin Transformer},'' \emph{Expert Systems with Applications}, vol. 260, p. 125437, 2025.

\bibitem{cv_task2}
D.-H. Hoang, A.-K. Tran, D.~N.~M. Dang, P.-N. Tran, H.~Dang-Ngoc, and C.~T. Nguyen, ``{RBBA: ResNet - BERT - Bahdanau Attention for Image Caption Generator},'' in \emph{2023 14th International Conference on Information and Communication Technology Convergence (ICTC)}, 2023, pp. 430--435.

\bibitem{medsam}
J.~Ma, Y.~He, F.~Li, L.~Han, C.~You, and B.~Wang, ``{Segment anything in medical images},'' \emph{Nature Communications}, vol.~15, no.~1, p. 654, 2024.

\bibitem{he2021transreid}
S.~He, H.~Luo, P.~Wang, F.~Wang, H.~Li, and W.~Jiang, ``{Transreid: Transformer-based object re-identification},'' in \emph{Proceedings of the IEEE/CVF international conference on computer vision}, 2021, pp. 15\,013--15\,022.

\bibitem{vit}
A.~Dosovitskiy, ``{An image is worth 16x16 words: Transformers for image recognition at scale},'' \emph{arXiv preprint arXiv:2010.11929}, 2020.

\bibitem{swin_transformer}
Z.~Liu \emph{et~al.}, ``{Swin transformer: Hierarchical vision transformer using shifted windows},'' in \emph{Proceedings of the IEEE/CVF international conference on computer vision}, 2021, pp. 10\,012--10\,022.

\bibitem{kaiming2016resnet}
K.~He, X.~Zhang, S.~Ren, and J.~Sun, ``Deep residual learning for image recognition,'' in \emph{2016 IEEE Conference on Computer Vision and Pattern Recognition (CVPR)}, 2016, pp. 770--778.

\bibitem{varghese2024yolov8}
R.~Varghese and S.~M., ``{YOLOv8: A Novel Object Detection Algorithm with Enhanced Performance and Robustness},'' in \emph{2024 International Conference on Advances in Data Engineering and Intelligent Computing Systems (ADICS)}, 2024, pp. 1--6.

\bibitem{isic2016}
D.~Gutman \emph{et~al.}, ``Skin lesion analysis toward melanoma detection: A challenge at the international symposium on biomedical imaging (isbi) 2016, hosted by the international skin imaging collaboration (isic),'' \emph{arXiv preprint arXiv:1605.01397}, 2016.

\bibitem{neopoly}
P.~Ngoc~Lan \emph{et~al.}, ``{NeoUNet : Towards Accurate Colon Polyp Segmentation and Neoplasm Detection},'' in \emph{Advances in Visual Computing}, G.~Bebis, V.~Athitsos, T.~Yan, M.~Lau, F.~Li, C.~Shi, X.~Yuan, C.~Mousas, and G.~Bruder, Eds.\hskip 1em plus 0.5em minus 0.4em\relax Cham: Springer International Publishing, 2021, pp. 15--28.

\bibitem{busi}
W.~Al-Dhabyani, M.~Gomaa, H.~Khaled, and A.~Fahmy, ``{Dataset of breast ultrasound images},'' \emph{Data in Brief}, vol.~28, p. 104863, 2020.

\bibitem{caicedo2019nucleus}
J.~C. Caicedo, A.~Goodman, K.~W. Karhohs, B.~A. Cimini, J.~Ackerman, M.~Haghighi, C.~Heng, T.~Becker, M.~Doan, C.~McQuin \emph{et~al.}, ``Nucleus segmentation across imaging experiments: the 2018 data science bowl,'' \emph{Nature methods}, vol.~16, no.~12, pp. 1247--1253, 2019.

\bibitem{jin2022fives}
K.~Jin, X.~Huang, J.~Zhou, Y.~Li, Y.~Yan, Y.~Sun, Q.~Zhang, Y.~Wang, and J.~Ye, ``Fives: A fundus image dataset for artificial intelligence based vessel segmentation,'' \emph{Scientific data}, vol.~9, no.~1, p. 475, 2022.

\bibitem{egeunet}
J.~Ruan, M.~Xie, J.~Gao, T.~Liu, and Y.~Fu, ``{EGE-UNet: An Efficient Group Enhanced UNet for Skin Lesion Segmentation},'' in \emph{Medical Image Computing and Computer Assisted Intervention -- MICCAI 2023}, H.~Greenspan, A.~Madabhushi, P.~Mousavi, S.~Salcudean, J.~Duncan, T.~Syeda-Mahmood, and R.~Taylor, Eds.\hskip 1em plus 0.5em minus 0.4em\relax Cham: Springer Nature Switzerland, 2023, pp. 481--490.

\bibitem{malunet}
J.~Ruan, S.~Xiang, M.~Xie, T.~Liu, and Y.~Fu, ``{MALUNet: A Multi-Attention and Light-weight UNet for Skin Lesion Segmentation},'' in \emph{2022 IEEE International Conference on Bioinformatics and Biomedicine (BIBM)}, 2022, pp. 1150--1156.

\bibitem{mhorunet}
R.~Wu \emph{et~al.}, ``{MHorUNet: High-order spatial interaction UNet for skin lesion segmentation},'' \emph{Biomedical Signal Processing and Control}, vol.~88, p. 105517, 2024.

\bibitem{vmunetv2}
M.~Zhang, Y.~Yu, S.~Jin, L.~Gu, T.~Ling, and X.~Tao, ``{VM-UNET-V2: Rethinking Vision Mamba UNet for Medical Image Segmentation},'' in \emph{Bioinformatics Research and Applications}, W.~Peng, Z.~Cai, and P.~Skums, Eds.\hskip 1em plus 0.5em minus 0.4em\relax Singapore: Springer Nature Singapore, 2024, pp. 335--346.

\bibitem{metnet}
A.~Iqbal and M.~Sharif, ``Memory-efficient transformer network with feature fusion for breast tumor segmentation and classification task,'' \emph{{Engineering Applications of Artificial Intelligence}}, vol. 127, p. 107292, 2024.

\bibitem{fscanet}
D.~Tan, R.~Hao, X.~Zhou, J.~Xia, Y.~Su, and C.~Zheng, ``{A Novel Skip-Connection Strategy by Fusing Spatial and Channel Wise Features for Multi-Region Medical Image Segmentation},'' \emph{IEEE Journal of Biomedical and Health Informatics}, vol.~28, no.~9, pp. 5396--5409, Sep. 2024.

\bibitem{zhao_hengshuang2017pspnet}
H.~Zhao, J.~Shi, X.~Qi, X.~Wang, and J.~Jia, ``Pyramid scene parsing network,'' in \emph{2017 IEEE Conference on Computer Vision and Pattern Recognition (CVPR)}, 2017, pp. 6230--6239.

\bibitem{badrinarayanan_vijay2017segnet}
V.~Badrinarayanan, A.~Kendall, and R.~Cipolla, ``Segnet: A deep convolutional encoder-decoder architecture for image segmentation,'' \emph{IEEE Transactions on Pattern Analysis and Machine Intelligence}, vol.~39, no.~12, pp. 2481--2495, 2017.

\bibitem{oktay2018attentionunet}
O.~Oktay, J.~Schlemper, L.~L. Folgoc, M.~Lee, M.~Heinrich, K.~Misawa, K.~Mori, S.~McDonagh, N.~Y. Hammerla, B.~Kainz \emph{et~al.}, ``Attention u-net: Learning where to look for the pancreas,'' \emph{arXiv preprint arXiv:1804.03999}, 2018.

\bibitem{liangchieh2018deeplabv3plus}
L.-C. Chen, Y.~Zhu, G.~Papandreou, F.~Schroff, and H.~Adam, ``{Encoder-Decoder with Atrous Separable Convolution for Semantic Image Segmentation},'' in \emph{Computer Vision -- ECCV 2018}, V.~Ferrari, M.~Hebert, C.~Sminchisescu, and Y.~Weiss, Eds.\hskip 1em plus 0.5em minus 0.4em\relax Cham: Springer International Publishing, 2018, pp. 833--851.

\bibitem{woo2018cbam}
S.~Woo, J.~Park, J.-Y. Lee, and I.~S. Kweon, ``Cbam: Convolutional block attention module,'' in \emph{Proceedings of the European conference on computer vision (ECCV)}, 2018, pp. 3--19.

\bibitem{xiang2019sk_cvpr}
X.~Li, W.~Wang, X.~Hu, and J.~Yang, ``Selective kernel networks,'' in \emph{2019 IEEE/CVF Conference on Computer Vision and Pattern Recognition (CVPR)}, 2019, pp. 510--519.

\bibitem{ji_lin2023sgaf}
J.~Lin, X.~Huang, H.~Zhou, Y.~Wang, and Q.~Zhang, ``Stimulus-guided adaptive transformer network for retinal blood vessel segmentation in fundus images,'' \emph{Medical Image Analysis}, vol.~89, p. 102929, 2023.

\bibitem{wu_kan2022tinyvit}
K.~Wu, J.~Zhang, H.~Peng, M.~Liu, B.~Xiao, J.~Fu, and L.~Yuan, ``Tinyvit: Fast pretraining distillation for small vision transformers,'' in \emph{Computer Vision -- ECCV 2022}, S.~Avidan, G.~Brostow, M.~Ciss{\'e}, G.~M. Farinella, and T.~Hassner, Eds.\hskip 1em plus 0.5em minus 0.4em\relax Cham: Springer Nature Switzerland, 2022, pp. 68--85.

\bibitem{mobilesam}
C.~Zhang, D.~Han, Y.~Qiao, J.~U. Kim, S.-H. Bae, S.~Lee, and C.~S. Hong, ``Faster segment anything: Towards lightweight sam for mobile applications,'' \emph{arXiv preprint arXiv:2306.14289}, 2023.

\end{thebibliography}
\end{document}